%% file: main.tex
\documentclass[lettersize,journal]{IEEEtran}
\usepackage{amsmath,amsfonts}
\usepackage{algorithmic}
\usepackage{algorithm}
\usepackage{array}
\usepackage[caption=false,font=normalsize,labelfont=sf,textfont=sf]{subfig}
\usepackage{textcomp}
\usepackage{stfloats}
\usepackage{url}
\usepackage{verbatim}
\usepackage{graphicx}
\usepackage{cite}
\usepackage{nth}
\usepackage{booktabs}
\usepackage{multirow}
\usepackage{amssymb}
\usepackage{array}
\usepackage{makecell}
\usepackage{colortbl}
\usepackage[table, dvipsnames]{xcolor}
\usepackage{amsmath}
\usepackage{dsfont}
\usepackage{hyperref}
\usepackage{threeparttable}
\usepackage{enumitem}
\usepackage[final,commandnameprefix=ifneeded]{changes} 

\setauthormarkup{}
\setdeletedmarkup{}
\setaddedmarkup{\textcolor{authorcolor}{#1}}

\newcommand{\gcell}[1]{\cellcolor{gray!15}#1}

\renewcommand{\footnoterule}{
    \kern -3pt
    \hrule width 0.4\columnwidth height 0.4pt
    \kern 2pt
}

\begin{document}

\title{\replaced{UniMM: A Unified Mixture Model Framework for Multi-Agent Simulation}{Revisit Mixture Models for Multi-Agent Simulation: Experimental Study within a Unified Framework}}

\author{Longzhong Lin, Xuewu Lin, Kechun Xu, Haojian Lu, Lichao Huang, Rong Xiong, Yue Wang\vspace{-20pt}
\thanks{Corresponding author: Yue Wang.}
\thanks{Longzhong Lin, Kechun Xu, Haojian Lu, Rong Xiong, and Yue Wang are with Zhejiang University, China (e-mail: \{\href{mailto:linlongzhong2000@zju.edu.cn}{linlongzhong2000}, \href{mailto:kcxu@zju.edu.cn}{kcxu}, \href{mailto:luhaojian@zju.edu.cn}{luhaojian}, \href{mailto:rxiong@zju.edu.cn}{rxiong}, \href{mailto:ywang24@zju.edu.cn}{ywang24}\}@zju.edu.cn).}
\thanks{Xuewu Lin, Lichao Huang are with Horizon Robotics, China (e-mail: \{\href{mailto:xuewu.lin@horizon.auto}{xuewu.lin}, \href{mailto:lichao.huang@horizon.auto}{lichao.huang}\}@horizon.auto).
This work was done during Longzhong Lin's internship at Horizon Robotics.}
\thanks{Project webpage: \href{https://longzhong-lin.github.io/unimm-webpage}{https://longzhong-lin.github.io/unimm-webpage}.}
\thanks{\copyright~2026 IEEE. Personal use of this material is permitted.  Permission from IEEE must be obtained for all other uses, in any current or future media, including reprinting/republishing this material for advertising or promotional purposes, creating new collective works, for resale or redistribution to servers or lists, or reuse of any copyrighted component of this work in other works.}
\thanks{This article has been published in IEEE Transactions on Pattern Analysis and Machine Intelligence. DOI: \href{https://doi.org/10.1109/TPAMI.2026.3700402}{10.1109/TPAMI.2026.3700402}.}}



\maketitle

\begin{abstract}
Simulation plays a crucial role in assessing autonomous driving systems, where the generation of realistic multi-agent behaviors is a key aspect.
In multi-agent simulation, the primary challenges include behavioral multimodality and closed-loop distributional shifts.
In this study, we \replaced{formulate a unified mixture model~(UniMM) framework}{revisit mixture models} for generating multimodal agent behaviors, which can cover the mainstream methods including \replaced{regression-based}{continuous} mixture models and \replaced{discrete NTP}{GPT-like discrete} models.
Furthermore, we introduce a closed-loop sample generation approach tailored for mixture models to mitigate distributional shifts.
Within the \deleted{unified mixture model~(}UniMM\deleted{)} framework, we recognize critical configurations from both \added{the} model and data perspectives.
We conduct a systematic examination of various model configurations, \replaced{and comprehensively characterize their effects}{including positive component matching, continuous regression, prediction horizon, and the number of components}.
Moreover, our investigation into the data configuration highlights the pivotal role of closed-loop samples in achieving realistic simulations.
To extend the benefits of closed-loop samples across a broader range of mixture models, we further \added{introduce a temporal disentanglement-and-alignment mechanism to} address the shortcut learning and off-policy learning issues.
Leveraging insights from our exploration, the distinct variants proposed within the UniMM framework, including discrete, anchor-free, and anchor-based models, all achieve state-of-the-art performance on the WOSAC benchmark.
\end{abstract}


\vspace{-9pt}
\section{Introduction}

\begin{figure*}[tb]
\centering
\includegraphics[width=0.97\textwidth]{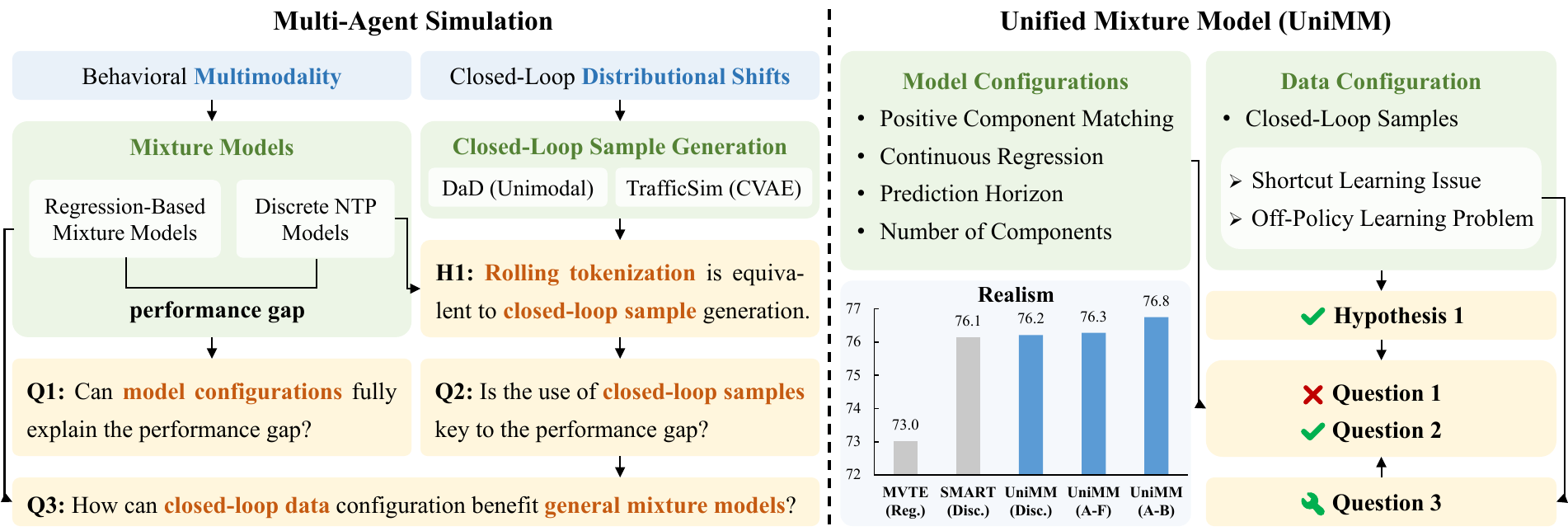}
\caption{\added{
The overview of our study.
For the main challenges in multi-agent simulation, specifically \textit{multimodality} and \textit{distributional shifts}, we \replaced{propose}{formulate} research questions centered on \textit{mixture models} and \textit{closed-loop sample generation}.
To investigate the hypothesis and questions, we \replaced{formulate}{revisit} the \textit{unified mixture model~(UniMM)} framework, recognizing critical configurations from both model and data perspectives.
Building on our exploration, the distinct variants proposed within the UniMM framework, including discrete, anchor-free, and anchor-based models, all achieve \textit{state-of-the-art} performance.
}}
\label{fig:overview}
\vspace{-11pt}
\end{figure*}

\IEEEPARstart{S}{imulation} facilitates the assessment of autonomous driving systems in a safe, controllable, and cost-effective manner.
One key to narrowing the gap between simulated and real-world environments is the generation of human-like multi-agent behaviors~\cite{suo2021trafficsim}.
In recent years, many studies~\cite{suo2021trafficsim,bergamini2021simnet,igl2022symphony,xu2023bits,zhong2023guided} have employed data-driven approaches to mimic the behavior of human traffic participants.
These imitative simulations typically learn a behavior model from real-world driving datasets, which generates future states for each agent based on map information and historical states.
Then the behavior model is iteratively run in an autoregressive manner to generate simulated scenarios.
The main challenges in achieving realistic simulation involve capturing the \textit{multimodality} of agent behaviors and addressing \textit{distributional shifts} in closed-loop rollouts.
Multimodality means that an agent may execute one of many underlying possible behaviors.
Distributional shifts refer to situations where the behavior model, when unrolled in a closed-loop manner, could encounter states rarely visited during training, thereby resulting in compounding errors.

Recovering the multimodality of agent behaviors from real-world datasets has been extensively studied in the field of motion prediction~\cite{gu2021densetnt, ngiam2021scene, girgis2021latent, zhou2022hivt, jia2023hdgt}.
State-of-the-art methods~\cite{shi2022motion, nayakanti2023wayformer, zhou2023query, lin2024eda, shi2024mtr++} predominantly employ mixture models, like Gaussian Mixture Models~(GMMs), to represent multimodal future motions.
Consequently, \replaced{several}{a bunch of} works~\cite{wang2023multiverse, qian20232nd, zhou2024behaviorgpt} have adopted analogous mixture models as behavior models for multi-agent simulation\added{, a family we refer to as \textit{regression-based mixture models}}.
These mixture models are usually trained to predict over a relatively long horizon, using a winner-takes-all continuous regression loss and a classification term based on the positive mixture component.
For identifying positive components, both anchor-based and anchor-free matching paradigms are commonly utilized.
More recently, inspired by the success of GPTs~\cite{floridi2020gpt, touvron2023llama}, a growing number of studies on multi-agent simulation~\cite{philion2024trajeglish, hu2025solving, wu2024smart, zhao2024kigras} have sought to discretize agent trajectories into motion tokens and apply a next-token prediction\added{~(NTP)} task\added{, yielding behavior models that we term \textit{discrete NTP models}.}
\replaced{These models predict}{where the behavior model predicts} a categorical distribution over a large number of motion modalities.
\replaced{
As shown in Table~\ref{table:survey}, the top-performing discrete NTP methods~\cite{wu2024smart, zhao2024kigras} achieve higher realism scores~\cite{montali2024waymo} than the strongest regression-based mixture models~\cite{zhou2024behaviorgpt, wang2023multiverse}, revealing an empirical \textit{performance gap} between these two lines of work.
}{
Overall, the GPT-like discrete models have empirically excelled in generating realistic simulations, outperforming continuous mixture models adapted from motion prediction.
}

\replaced{
In this work, we adopt a unified view that treats both regression-based mixture models and discrete NTP models as instances of a unified mixture model~(\textbf{UniMM}) framework.
Under this view, a categorical distribution over motion tokens can be interpreted as a mixture distribution, where each component represents a discrete category and each motion token serves as an anchor for its corresponding component.
Accordingly, in this unified framework, the leading discrete NTP methods~\cite{wu2024smart, zhao2024kigras} can be instantiated as anchor-based mixture models without continuous regression~(Section~\ref{Continuous Regression}), typically with a large number of components and a short prediction horizon.}{
In fact, we note that GPT-like models with categorical distributions can also be viewed as mixture models, where each mixture component represents a discrete category.
The motion tokens, in this context, are analogous to anchors, each linked to a specific mixture component.
As mixture models, the leading GPT-like methods~\cite{wu2024smart, zhao2024kigras} are essentially anchor-based models devoid of continuous regression, typically with a large number of components and a short prediction horizon.
}
Therefore, the distinction in model design between \replaced{regression-based mixture models and discrete NTP models}{the aforementioned discrete and continuous models} can be understood as different model configuration choices within \replaced{the UniMM framework}{the framework of mixture models}.
Given this perspective, we wonder \textbf{Q1:} \textit{Can the differences in model configurations fully account for the performance gap} between existing \replaced{regression-based mixture models and discrete NTP models}{continuous and discrete mixture models}?

In addition to behavioral multimodality, another major challenge of multi-agent simulation lies in distributional shifts during closed-loop rollouts.
In time series modeling, DaD~\cite{venkatraman2015improving} mitigates distributional shifts by leveraging model predictions to modify ground truth inputs in training samples.
TrafficSim~\cite{suo2021trafficsim} employs an analogous principle for multi-agent simulation with a CVAE-based behavior model, where ground truth states are autoregressively replaced with posterior predictions from the CVAE, forming closed-loop samples.
To discretize ground truth trajectories, \replaced{discrete NTP}{GPT-like} methods also transform the training samples, via a tokenization process.
This motion tokenization often applies a rolling matching strategy to reduce discretization errors across the full sequence, where the continuous states are iteratively substituted by matched motion tokens~\cite{philion2024trajeglish, wu2024smart}.
The above methods all infuse training samples with model-related features through rollouts.
Observing this similarity in data processing, we hypothesize \textbf{H1:} \textit{Motion tokenization with rolling matching is equivalent to the generation of closed-loop samples}.
If the hypothesis holds, both being mixture models, \replaced{discrete NTP}{GPT-like discrete} methods can be seen as adopting a closed-loop data configuration, while existing \replaced{regression-based mixture}{continuous} models directly utilize the original data.
This leads us to ask \textbf{Q2:} \textit{Is the data configuration involving closed-loop samples responsible for the performance gap} between \replaced{existing regression-based mixture models and discrete NTP models}{continuous and discrete mixture models}?
Moreover, we inquire \textbf{Q3:} \textit{How might such a data configuration be extended to benefit general mixture models}?

As shown in Fig.~\ref{fig:overview}, to investigate the aforementioned hypothesis and questions,
\replaced{
we conduct a systematic empirical study within the UniMM framework from both the model and data perspectives.
On the model side~(\textbf{Q1}), we identify the critical configuration dimensions and comprehensively characterize their effects.
On the data side~(\textbf{Q2}), we propose a closed-loop sample generation approach for general mixture models, and clarify how this mechanism manifests in discrete NTP models~(\textbf{H1}).
To combine advantageous configurations from both the model and data sides~(\textbf{Q3}), we uncover two key issues that prevent closed-loop samples from benefiting models with long prediction horizons: shortcut learning and off-policy learning.
To address these issues, we propose a temporal disentanglement-and-alignment mechanism for data generation and model training, thereby integrating the strengths of closed-loop data and long-horizon prediction.
Additionally, by leveraging predefined anchors in both architecture and data collection, we enable anchor-based models with continuous regression to attain efficiency comparable to their discrete counterparts.
Ultimately, our UniMM-derived solutions advance regression-based mixture models---across both anchor-free and anchor-based settings---to state-of-the-art performance on the WOSAC benchmark~\cite{montali2024waymo}, effectively narrowing and even reversing the performance gap between these two model families.
Beyond public benchmarks, we further integrate UniMM into the MetaDrive~\cite{li2022metadrive} simulator to support driving-policy evaluation, demonstrating its practicality as a simulator-ready background traffic model for real-world–oriented applications.
}{
we review the general paradigm of mixture models for multi-agent simulation, seeking to discuss methods derived from different inspirations within a unified framework.
Subsequently, we highlight several critical model configurations, including positive component matching, continuous regression, prediction horizon, and number of components.
Benefiting from the unified framework, we systematically examine mixture models across these diverse configurations.
Furthermore, to explore the data configuration involving closed-loop samples, we propose a closed-loop sample generation approach for general mixture models, drawing on the philosophy of DaD~\cite{venkatraman2015improving} and TrafficSim~\cite{suo2021trafficsim}.
When it comes to \replaced{discrete NTP}{GPT-like} models, namely anchor-based models without continuous regression, we \replaced{show in Section~\ref{Closed-Loop Sample Generation}}{confirm \textbf{H1}, showing} that the above closed-loop sample generation is equivalent to motion tokenization with rolling matching\added{, as hypothesized in \textbf{H1}}.
We thus identify a commonly overlooked disparity between existing \replaced{regression-based mixture models and discrete NTP models}{continuous and discrete mixture models}: discrete \added{NTP} models naturally train on closed-loop samples introduced by tokenization, while their \replaced{regression-based}{continuous} counterparts do not.
}

\added{
Our main contributions are as follows:
(1)~For multi-agent simulation, we formulate a unified mixture model~(UniMM) framework that bridges mainstream methods originating from distinct modeling paradigms, and identify the critical configurations from both the model and data perspectives.
(2)~We present the hypothesis \textbf{H1} and the research questions \textbf{Q1--Q3}, and investigate them through a systematic empirical study within the UniMM framework.
(3)~We propose a closed-loop sample generation approach for general mixture models, identify the practical challenges in combining the benefits of closed-loop samples with long prediction horizons, and introduce a temporal disentanglement-and-alignment mechanism to address these challenges.
(4)~Our UniMM-derived designs enable regression-based mixture models---across both anchor-free and anchor-based settings---to match or exceed the performance of discrete NTP models, achieving state-of-the-art results on WOSAC~\cite{montali2024waymo}.
(5)~We integrate UniMM into MetaDrive~\cite{li2022metadrive} for simulation-based evaluation, demonstrating its practicality for real-world–oriented applications.
}

\section{Related Work}
\subsection{Multi-Agent Simulation}
Simulation provides a safe, controllable, and cost-effective environment for assessing autonomous driving systems.
Earlier simulators~\cite{pomerleau1988alvinn, krajzewicz2002sumo, dosovitskiy2017carla, lopez2018microscopic} utilized heuristic-based policies~\cite{gipps1981behavioural, kesting2007general, treiber2000congested} to model agent behaviors, which often struggle to capture the complexity of real-world scenarios.
Benefiting from the availability of real-world driving datasets~\cite{caesar2020nuscenes, ettinger2021large}, learning-based imitative methods~\cite{suo2021trafficsim,bergamini2021simnet,igl2022symphony,xu2023bits,zhong2023guided} have demonstrated superiority in generating human-like multi-agent simulations.
The Waymo Open Sim Agents Challenge~(WOSAC)~\cite{montali2024waymo}, based on the large-scale Waymo Open Motion Dataset~(WOMD)~\cite{ettinger2021large}, is the current mainstream benchmark for multi-agent simulation, evaluating the realism of simulated scenarios through their consistency with real-world data distributions.
In this study, we perform experiments on the WOMD and adopt the WOSAC metrics for evaluation.

\subsection{Modeling for Multimodal Agent Behavior}
Modeling multimodal agent behaviors has been widely studied in motion prediction, as this task also requires predicting multiple future trajectories of an agent within the same context.
In recent years, various kinds of generative models, including VAEs~\cite{lee2017desire,suo2021trafficsim,rempe2022generating}, GANs~\cite{gupta2018social,rhinehart2018r2p2,igl2022symphony}, and diffusion models~\cite{jiang2023motiondiffuser,zhong2023guided,huang2024versatile}, have been investigated for both motion prediction and multi-agent simulation.

In the field of motion prediction, state-of-the-art methods~\cite{shi2022motion, nayakanti2023wayformer, zhou2023query, lin2024eda, shi2024mtr++} predominantly utilize mixture models to represent multimodal future motions.
These mixture models typically employ a winner-takes-all continuous regression loss along with a classification term, computed on the positive mixture component.
For selecting positive components, the mixture models mainly fall into two categories: anchor-free and anchor-based.
Anchor-free models~\cite{tang2019multiple,varadarajan2022multipath++,nayakanti2023wayformer} select the positive component by directly comparing predicted trajectories to the ground truth, featuring flexible predictions for each component.
In contrast, anchor-based models~\cite{chai2019multipath,zhao2021tnt,gilles2021home,gilles2022gohome,shi2024mtr++} associate each component with an anchor and select the positive one matching the closest anchor to ground truth.
Due to parallels in functional forms, \replaced{several}{a bunch of} works on multi-agent simulation have adopted \replaced{regression-based}{continuous} mixture models similar to those in motion prediction, including both anchor-free~\cite{zhou2024behaviorgpt} and anchor-based~\cite{wang2023multiverse,qian20232nd} models.
These approaches usually predict over a relatively long horizon to facilitate long-range interaction reasoning.

Inspired by GPTs~\cite{floridi2020gpt, touvron2023llama}, an increasing number of recent studies on multi-agent simulation~\cite{philion2024trajeglish, hu2025solving, wu2024smart, zhao2024kigras} have adopted next-token prediction by discretizing trajectories into motion tokens.
These \replaced{discrete NTP}{GPT-like discrete} models commonly question the ability of \replaced{regression-based}{continuous} mixture models with limited number of components to capture behavioral multimodality.
Thus, they typically predict a categorical distribution over a large set of motion tokens and have indeed demonstrated superior performance.
In this study, we observe that \replaced{discrete NTP models}{GPT-like models} can also be considered mixture models, with the differentiation between the above discrete and continuous models being interpreted as differing configuration choices within the unified mixture model framework.

\subsection{Closed-Loop Samples for Distributional Shifts}
To alleviate distributional shifts during closed-loop rollouts, DaD~\cite{venkatraman2015improving} for time series modeling proposes substituting ground truth inputs in training samples with autoregressive model predictions, which inherits theoretical guarantees from the online learning algorithm DAgger~\cite{ross2011reduction}.
Nevertheless, DaD focuses solely on the case of unimodal single-step prediction models.
TrafficSim~\cite{suo2021trafficsim} leverages an akin principle for multi-agent simulation with a CVAE-based behavior model.
Specifically, it unrolls with predictions decoded from latent variables of the posterior encoder, which incorporates ground truth outputs, to generate closed-loop samples for training.
\replaced{Discrete NTP}{GPT-like} methods often employ rolling matching~\cite{philion2024trajeglish,wu2024smart} during motion tokenization, autoregressively replacing continuous states with matched motion tokens, a process similar to the data processing in TrafficSim but aimed at reducing discretization errors.
In this study, we propose a closed-loop sample generation approach tailored for general mixture models, and pinpoint its connection to motion tokenization with rolling matching in \replaced{discrete NTP}{GPT-like} models.

\section{Problem Formulation} \label{Problem Formulation}

\subsection{Multi-Agent Simulation}
Generally, the generation of multi-agent simulations is factorized into an autoregressive sequential process:
\footnote{
For simplicity, we use $S_{t_1:t_2}$ to represent the state trajectory over the time interval $(t_1, t_2]$.
In practice, state trajectories are typically parameterized by discrete sampling points.
For example, $S_{0:0.5s}$ can be expressed as a state sequence sampled at 10 Hz:
$S_{0:0.5s} = \{S_{0.1s}, S_{0.2s}, S_{0.3s}, S_{0.4s}, S_{0.5s}\}$.
}
\begin{equation}
p(S_{0:T}|C_0) = \prod_{t\in\{0,\tau,2\tau,\dots,T-\tau\}} p(S_{t:t+\tau}|S_{0:t},C_0),
\end{equation}
where $S_{0:T}$ represents the simulated scenario, specifically the state sequence of all agents from time $0$ to $T$, with $T$ being the total simulation duration.
$C_0$ denotes the initial scenario context, including map information and historical agent states before time $0$.
The agent state typically includes the agent's position, heading, and other attributes.
$\tau$ represents the update interval of the simulation.

With a sufficiently small update interval $\tau$, each agent can be considered to independently take actions, given adequate historical context~\cite{seff2023motionlm,philion2024trajeglish}:
\begin{equation}
p(S_{t:t+\tau}|S_{0:t},C_0) = \prod_{n=1}^{N} p(S_{t:t+\tau}^{n}|S_{0:t},C_0,n),
\end{equation}
where $S_{t:t+\tau}^{n}$ represents the state sequence of the $n$-th agent from time $t$ to $t+\tau$, and $N$ denotes the number of agents in the scenario.
In practice, we adopt the widely used setting of $\tau=0.5s$~\cite{wu2024smart,zhao2024kigras}, which aligns with real-world driving~\cite{engstrom2024modeling}.

\subsection{Mixture Model for Behavioral Multimodality}
To generate realistic multi-agent simulations, data-driven approaches typically learn a behavior model:
\begin{equation}
\pi_{\theta}(Y|X), \ \text{where}
\left\{
\begin{aligned}
X &:= (S_{0:t}, C_0, n) \\
Y &:= S_{t:t+T_{\text{pred}}}^{n} \\
\end{aligned}
\right.
.
\label{eq:prediction_horizon}
\end{equation}
The behavior model $\pi_{\theta}$ predicts $Y$, the state trajectory of the $n$-th agent over the prediction horizon $T_{\text{pred}}$, given $X$ which includes the historical agent states, the initial scenario context, and the agent identifier.
For modeling multimodal agent behaviors, \textbf{mixture models} are a suitable choice~\cite{shi2024mtr++}:
\begin{equation}
\pi_{\theta}(Y|X) = \sum_{k=1}^{K} q_{\theta}(Z=k|X) m_{\theta}(Y|Z=k,X),
\label{eq:component_number}
\end{equation}
where $K$ is the number of mixture components, and $Z$ is the latent variable indicating the component selection.
$q_{\theta}(Z|X)$ predicts the probabilities of selecting each component, while $m_{\theta}(Y|Z,X)$ represents the distribution of the future trajectory conditioned on the selected component.
For instance, in a Gaussian Mixture Model~(GMM), the component distribution $m_{\theta}$ corresponds to a unimodal Gaussian distribution.

\subsubsection{Imitative Training}
Given the dataset $\mathcal{D}$ of real-world driving scenarios, the mixture model $\pi_{\theta}$ is optimized through maximum likelihood estimation~(MLE):
\begin{equation}
\max_{\theta} \mathbb{E}_{(x,y) \sim \mathcal{D}}[\log \pi_{\theta}(y|x)], \ \text{where}
\left\{
\begin{aligned}
x &:= (s_{0:t},c_0,n) \\
y &:= s_{t:t+T_{\text{pred}}}^{n} \\
\end{aligned}
\right.
,
\label{eq:open-loop_samples}
\end{equation}
where $c_0$, $s_{0:t}$, $s_{t:t+T_{\text{pred}}}^{n}$ are respectively the sampled values of $C_0$, $S_{0:t}$, and $S_{t:t+T_{\text{pred}}}^{n}$ from the dataset $\mathcal{D}$.
The above log-likelihood can be decomposed into the sum of the evidence lower bound~(ELBO) and the KL-divergence between the posterior distribution and an arbitrary distribution $q$ over the latent variable $z$~\cite{neal1998view}:
\begin{equation}
\begin{aligned}
\log \pi_{\theta}(y|x) =& \underbrace{\mathbb{E}_{z \sim q}[\log m_\theta(y|z,x)] - \mathbb{KL}[q(z)\|q_\theta(z|x)]}_{ELBO} \\
&+ \mathbb{KL}[q(z) \| p(z|y,x;\theta)].
\end{aligned}
\end{equation}
Thus, according to the EM algorithm~\cite{dempster1977maximum}, the optimization objective can be reformulated as:
\begin{equation}
\begin{aligned}
&\max_{\theta} \mathbb{E}_{(x,y) \sim \mathcal{D}}\Big[\mathbb{E}_{z \sim q^*}[\log m_\theta(y|z,x)] - \mathbb{KL}[q^*(z) \| q_\theta(z|x)]\Big],\\
& \ \text{where} \ q^*(z) = p(z|y,x;\theta) = \frac{q_\theta(z|x)m_\theta(y|z,x)}{\sum_{z} \ q_\theta(z|x)m_\theta(y|z,x)}.
\end{aligned}
\end{equation}
In practice, a hard-assignment strategy is adopted to approximate the true posterior~\cite{chai2019multipath,varadarajan2022multipath++}:
\begin{equation}
q^*(z) \approx \hat{q}^*(z) := \mathds{1}(z=z^*),
\label{eq:positive_component}
\end{equation}
where $\mathds{1}(\cdot)$ denotes the indicator function, and $z^*$ represents the positive component of the mixture model $\pi_{\theta}$ that best matches the ground truth $(x, y)$.
Therefore, the final training objective of the mixture model is as follows, incorporating a winner-takes-all regression loss and a classification term:
\begin{equation}
\max_{\theta} \mathbb{E}_{(x,y) \sim \mathcal{D}}\Big[\underbrace{\log m_\theta(y|z^*,x)}_{\text{regression}} - \underbrace{\mathbb{KL}[\hat{q}^*(z) \| q_\theta(z|x)]}_{\text{classification}}\Big].
\label{eq:training_objective}
\end{equation}

\subsubsection{Model Configurations}
Within the unified framework of mixture models that generate multimodal agent behaviors, we elaborate on the critical model configurations in Section~\ref{Model Configurations}, which include\deleted{:}
\added{(1)}~\textbf{\replaced{prediction}{Prediction} horizon} $T_{\text{pred}}$ in Eq.~\ref{eq:prediction_horizon}\replaced{;}{.}
\added{(2)}~\textbf{\replaced{num-ber}{Number} of components} $K$ in Eq.~\ref{eq:component_number}\replaced{;}{.}
\added{(3)}~\textbf{\replaced{positive}{Positive} component matching} for $z^*$ in Eq.~\ref{eq:positive_component}\replaced{;}{.}
\added{(4)}~\textbf{\replaced{continuous}{Continuous} regression} associated with $m_\theta$ in Eq.~\ref{eq:training_objective}.

\subsection{Distributional Shifts in Closed-Loop Simulation}

\begin{figure}[tb]
\centering
\includegraphics[width=0.9\columnwidth]{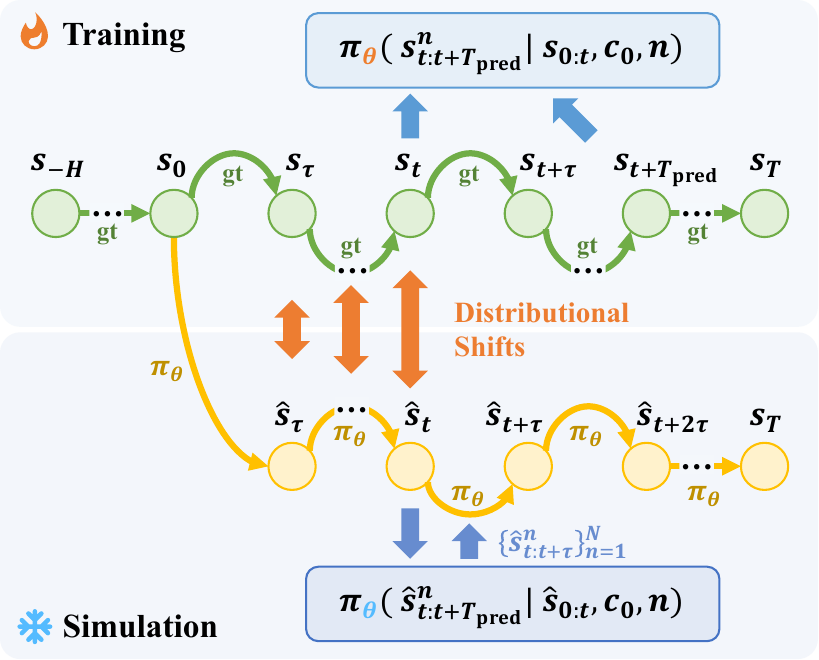}
\caption{
The demonstration of the imitative training on \textit{open-loop} samples and the \textit{closed-loop} simulation process for the behavior model $\pi_\theta$.
}
\label{fig:training_simulation}
\vspace{-11pt}
\end{figure}

As shown in Fig.~\ref{fig:training_simulation}, if the behavior model $\pi_{\theta}$ is naively trained on \textit{open-loop samples}, directly derived from splits of ground truth trajectories $(x=(s_{0:t},c_0,n), y) \sim \mathcal{D}$, it could be prone to distributional shifts and suffer from compounding errors in closed-loop simulation~\cite{ross2011reduction}.
Specifically, when $\pi_{\theta}$ is unrolled autoregressively during simulation:
\begin{equation}
\begin{aligned}
\hat{s}^n_{t:t+T_{\text{pred}}} &\sim \pi_{\theta}(\hat{s}^n_{t:t+T_{\text{pred}}}|\hat{s}_{0:t},c_0,n), \\
\hat{s}_{t:t+\tau} &= \{\hat{s}^n_{t:t+\tau}\}_{n=1}^N,
\end{aligned}
\end{equation}
the model may encounter novel input states $\hat{s}_{0:t}$ induced by its previous suboptimal predictions, which differ from the observed states $s_{0:t}$ during training.
As the input states $\hat{s}_{0:t}$ deviate from the training data, the prediction errors increase, which in turn exacerbates the distributional shift of the subsequent input states $\hat{s}_{0:t+\tau}$, leading to compounding errors.

\subsubsection{Data Configuration}
As proposed in DaD~\cite{venkatraman2015improving}, utilizing closed-loop samples $(x^{\text{cl}}, y) \sim \mathcal{D}^{\text{cl}}$ can effectively mitigate distributional shifts:
\begin{equation}
\max_{\theta} \mathbb{E}_{(x^{\text{cl}},y) \sim \mathcal{D}^{\text{cl}}}[\log \pi_{\theta}(y|x^{\text{cl}})],
\label{eq:closed-loop_samples}
\end{equation}
\replaced{$\mathcal{D}^{\text{cl}}$ denotes the distribution of closed-loop samples derived from the original dataset $\mathcal{D}$, where the ground truth input $x$ in $(x, y) \sim \mathcal{D}$ is replaced by the corresponding closed-loop input $x^{\text{cl}}$ generated via autoregressive predictions, while the ground truth output $y$ remains unchanged.}{where the ground truth input $x$ in Eq.~\ref{eq:open-loop_samples} is substituted with the closed-loop input $x^{\text{cl}}$, generated from the corresponding autoregressive predictions.}
However, DaD~\cite{venkatraman2015improving} focuses solely on unimodal, single-step prediction models.
To extend this principle to mixture models for multimodal and long-horizon behavior modeling, we further investigate the analogous data configuration within the unified mixture model framework in Section~\ref{Data Configuration}, centering on\deleted{:}
\textbf{\replaced{closed-loop}{Closed-loop} samples} $(x^{\text{cl}}, y) \sim \mathcal{D}^{\text{cl}}$ in Eq.~\ref{eq:closed-loop_samples}, derived by transforming open-loop samples $(x, y) \sim \mathcal{D}$.


\begin{table*}[htb]
\centering
\caption{
\added{Configurations to Explore and WOSAC Realism Meta Metrics of Recent SOTA Methods Based on Mixture Models}
}
\resizebox{0.99\textwidth}{!}{
\def\arraystretch{1.3}
\begin{threeparttable}
\begin{tabular}{c|l|ccccc|cc}
\specialrule{1pt}{0pt}{0pt}
\multirow{2}{*}{Type} & \multirow{2}{*}{Method} & \multirow{2}{*}{\makecell{Anchor-Based\\Matching}} & \multirow{2}{*}{\makecell{Continuous\\Regression}} & \multirow{2}{*}{\makecell{Prediction\\Horizon}} & \multirow{2}{*}{\makecell{Number of\\ Components}} & \multirow{2}{*}{\makecell{Closed-Loop\\Samples}} & \multirow{2}{*}{\makecell{Realism $\uparrow$\\ \href{https://waymo.com/open/challenges/2023/sim-agents/}{(2023)}}} & \multirow{2}{*}{\makecell{Realism $\uparrow$\\ \href{https://waymo.com/open/challenges/2024/sim-agents/}{(2024)}}} \\ & & & & & & & \\
\hline
\multirow{4}{*}{\makecell{Discrete\\NTP}}
& SMART~\cite{wu2024smart} & $\checkmark$ & $\times$ & 0.5s & 1024 & $\checkmark$ & \textbf{0.6587} & \textbf{0.7591} \\
& KiGRAS~\cite{zhao2024kigras} & $\checkmark$ & $\times$ & 0.5s & 3969 & -\tnote{*} & - & \textbf{0.7597} \\
& GUMP~\cite{hu2025solving} & $\checkmark$ & $\times$ & 2s & $1.6\times10^{33}$\tnote{\dag} & $\checkmark$ & 0.6432 & 0.7431 \\
& Trajeglish~\cite{philion2024trajeglish} & $\checkmark$ & $\times$ & 0.1s & 384 & $\checkmark$ & 0.6451 & - \\
\hline
\multirow{2}{*}{\makecell{Regression-\\Based}}
& BehaviorGPT~\cite{zhou2024behaviorgpt} & $\times$ & $\checkmark$ & 1s & 16 & $\times$ & - & \textbf{0.7473} \\
& MVTE~\cite{wang2023multiverse} & $\checkmark$ & $\checkmark$ & 1s & 64 & $\times$ & \textbf{0.6448} & 0.7302 \\
\specialrule{1pt}{0pt}{0pt}
\end{tabular}
\begin{tablenotes}[flushleft]
\scriptsize
\item[\dag] As the GUMP directly utilizes grid-based discretization of scene-centric coordinates, each 0.5s corresponds to $2\times10^8$ tokens, resulting in an extremely large $(2\times10^8)^4$ components over a 2-second period.
\item[*] KiGRAS generates samples containing closed-loop states during the agent motion tokenization process, but it is unclear whether the model utilizes these samples for training.
\end{tablenotes}
\end{threeparttable}
}
\label{table:survey}
\vspace{-11pt}
\end{table*}

\section{Model Configurations} \label{Model Configurations}
\subsection{Positive Component Matching} \label{Positive Component Matching}
For selecting the positive component $z^*$ in Eq.~\ref{eq:positive_component}, mixture models can be categorized into two primary paradigms: \textbf{anchor-free} and \textbf{anchor-based} matching.

\begin{figure}[tb]
\centering
\includegraphics[width=0.99\columnwidth]{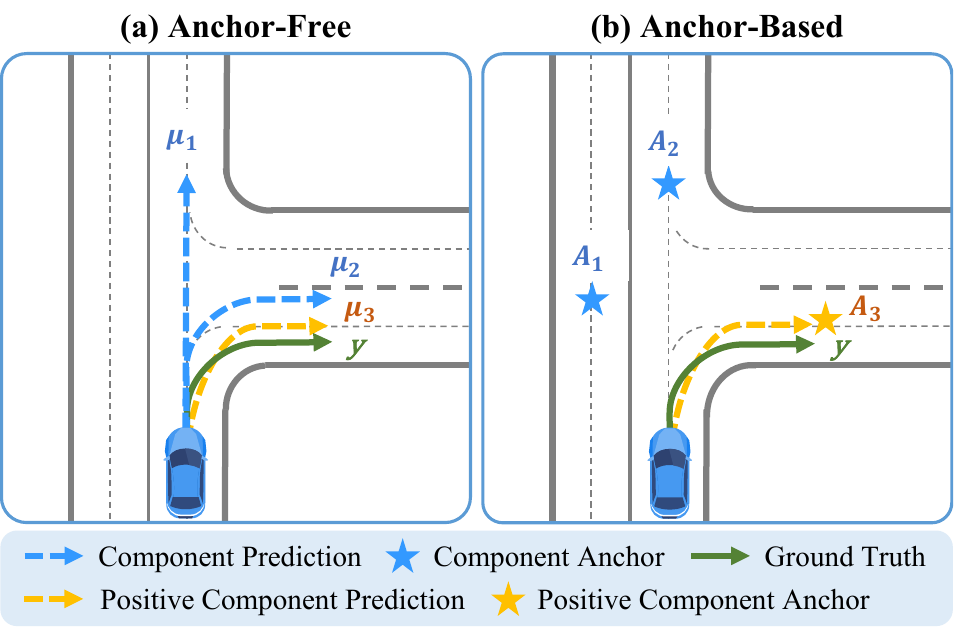}
\caption{
The two primary positive component matching paradigms.
(a) and (b) respectively present the \textit{anchor-free} and \textit{anchor-based} matching.
}
\label{fig:positive_component}
\vspace{-11pt}
\end{figure}

\subsubsection{Anchor-Free Matching}
As demonstrated in Fig.~\ref{fig:positive_component}(a), anchor-free methods~\cite{nayakanti2023wayformer,zhou2024behaviorgpt} designate the component $z^*$, whose predicted trajectory $\mu_{z^*}(x;\theta)$ is closest to the ground truth $y$, as positive:
\begin{equation}
z^* = \arg\min_{k} d\big(\mu_k(x;\theta),y\big),
\label{eq:anchor-free}
\end{equation}
where $d(\cdot,\cdot)$ computes the distance between two trajectories, and $\mu_k(x;\theta) := \mathbb{E}_{Y \sim m_{\theta}(Y|Z=k,X=x)}[Y]$ represents the predicted trajectory corresponding to component $k$.

\subsubsection{Anchor-Based Matching}
In anchor-based models~\cite{shi2024mtr++,wang2023multiverse,philion2024trajeglish}, the state anchors $\{A_k(x)\}_{k=1}^{K}$ are each associated with a specific component.
As illustrated in Fig.~\ref{fig:positive_component}(b), the positive component is determined as the one corresponding to the anchor closest to ground truth:
\begin{equation}
z^* = \arg\min_{k} d\big(A_k(x),y\big).
\label{eq:anchor-based}
\end{equation}
$A_k(x)$ means that anchors can be predefined in spaces like 
agent-centric trajectories~\cite{wu2024smart}, and then transformed into the same state space as the ground truth $y$ using state information in $x$.
Additionally, the distribution of anchors can be influenced by information such as the agent category~\cite{shi2022motion} in $x$.

\subsubsection{Anchor-Free vs. Anchor-Based}
According to prior research on motion prediction~\cite{lin2024eda}, mixture models employing different matching paradigms exhibit distinct tendencies.
Anchor-free models, due to their flexible predictions for each component, make it hard for $q_\theta$ to distinguish the component closest to the ground truth.
Thus, their performance relies more heavily on the regression capability of $m_\theta$.
In anchor-based models, the introduction of stably distributed anchors significantly alleviates the difficulty for $q_\theta$ in selecting the correct positive component, and these models tend to generate trajectories around the anchors.
So their effectiveness is more dependent on the classification performance of $q_\theta$.

In this study, we implement both anchor-free and anchor-based models under relatively consistent and fair conditions, enabling a direct demonstration of the characteristic differences between the two matching paradigms in the context of multi-agent simulation.

\subsection{Continuous Regression} \label{Continuous Regression}
For anchor-based mixture models described in Section~\ref{Positive Component Matching}, if the component distribution $m_\theta$ is chosen as:
\begin{align}
m_\theta(Y|Z=k,X=x) = \mathds{1}(Y=A_k(x)),
\end{align}
which means the state anchor $A_k(x)$ is directly used as the predicted trajectory for component $k$.
At this point, the component distribution is no longer dependent on trainable parameters $\theta$.
Hence, the regression term in the training objective~(Eq.~\ref{eq:training_objective}) is redundant, leaving only the classification loss:
\begin{equation}
\begin{aligned}
&\max_{\theta} \mathbb{E}_{(x,y) \sim \mathcal{D}}\Big[- \mathbb{KL}[\hat{q}^*(z) \| q_\theta(z|x)]\Big] \\
\Leftrightarrow &\min_{\theta} \mathbb{E}_{(x,y) \sim \mathcal{D}}\Big[\mathcal{L}_{\text{CE}}\big(\mathds{1}(z=z^*),q_\theta(z|x)\big)\Big],
\end{aligned}
\end{equation}
where $\mathcal{L}_{\text{CE}}$ denotes the cross-entropy loss.

The above setting fully aligns with \replaced{discrete NTP models~\cite{wu2024smart, zhao2024kigras, hu2025solving, philion2024trajeglish}}{GPT-like models with categorical distributions~\cite{wu2024smart, zhao2024kigras, hu2025solving, philion2024trajeglish}}.
Thus, \textit{\replaced{discrete NTP}{GPT-like discrete} models are essentially anchor-based mixture models, devoid of the trainable component distribution and its corresponding continuous regression}.
Among previous works, the discrete models often outperform those with continuous regression.
However, as shown in Table~\ref{table:survey}, these methods also exhibit notable differences in other configurations.

In our study, we evaluate anchor-based models, both with and without continuous regression, under consistent configurations aligned with the leading discrete models, to validate whether incorporating continuous regression provides a performance advantage.

\subsection{Prediction Horizon}
As indicated in previous works~\cite{suo2021trafficsim,hu2025solving,zhou2024behaviorgpt}, training behavior models with a longer prediction horizon, $T_{\text{pred}}$~(Eq.~\ref{eq:prediction_horizon}), may enhance spatio-temporal interaction reasoning, helping agents become more robust to distributional shifts and generate realistic behaviors.
However, these studies merely provide a preliminary demonstration of the benefits associated with increasing the prediction horizon.

In our experiments, by systematically examining a broad range of $T_{\text{pred}}$ under diverse conditions, we observe a more comprehensive trend in the variation of simulation realism as the prediction horizon increases, along with the interactions between the prediction horizon and other configurations.

\subsection{Number of Components}
Intuitively, the number of components, $K$~(Eq.~\ref{eq:component_number}), reflects the mixture model's ability to represent complex distributions.
Some prior works~\cite{zhou2024behaviorgpt} have also demonstrated the positive impact of using more mixture components.
As shown in Table~\ref{table:survey}, discrete models typically have a much larger number of components, which is commonly considered an important factor behind their superior performance~\cite{philion2024trajeglish, wu2024smart}.

In the experiments, we systematically investigate mixture models with varying component numbers.
Through the related model architecture design, which will be discussed in Section~\ref{Motion Decoder}, anchor-based mixture models with continuous regression are enabled to scale up the number of components $K$ as efficiently as those employing categorical distributions.
It is observed that anchor-based models continue to benefit from the growth of $K$, whereas the situation differs for anchor-free models.
Notably, we find that anchor-free models with 6 components can also achieve highly competitive performance, based on the use of closed-loop data samples to be introduced in Section~\ref{Data Configuration}.

\section{Data Configuration} \label{Data Configuration}
This section focuses on the utilization of \textbf{closed-loop samples}.
We begin by introducing the closed-loop sample generation method proposed for general mixture models~(Section~\ref{Closed-Loop Sample Generation}).
Next, we discuss the potential challenges\deleted{ associated with training on closed-loop samples,} including the shortcut learning issue~(Section~\ref{Shortcut Learning Issue}) and the off-policy learning problem~(Section~\ref{Off-Policy Learning Problem}).
\added{We then present a temporal disentanglement-and-alignment mechanism~(Section~\ref{Disentanglement and Alignment of Horizons}) to address these issues.}
Finally, we propose an approximate \replaced{closed-loop data generation}{posterior} policy~(Section~\ref{Approximate Posterior Policy}) to accelerate the closed-loop sample generation for anchor-based models.

\subsection{Closed-Loop Sample Generation} \label{Closed-Loop Sample Generation}

\begin{figure}[tb]
\centering
\includegraphics[width=0.85\columnwidth]{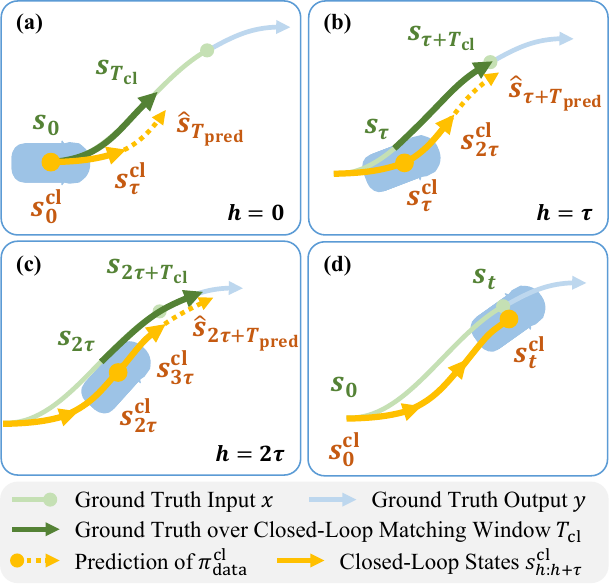}
\caption{\added{
The illustration of closed-loop sample generation.
(a), (b), (c), and (d) sequentially represent the steps of the process.
}}
\label{fig:closed-loop_samples}
\vspace{-11pt}
\end{figure}


Inspired by DaD~\cite{venkatraman2015improving} and TrafficSim~\cite{suo2021trafficsim}, we propose a closed-loop sample generation method tailored for mixture models to mitigate distributional shifts.
\replaced{
As illustrated in Fig.~\ref{fig:closed-loop_samples}, given an open-loop sample $(x = (s_{0:t}, c_0, n), y)$ from the dataset $\mathcal{D}$~(Eq.~\ref{eq:open-loop_samples}), we progressively transform the ground truth input states $s_{0:t}$ into the closed-loop input states $s^{\text{cl}}_{0:t}$ that incorporate the predicted behaviors of $\pi_\theta$, thereby obtaining a closed-loop sample:
}{
Briefly, given an open-loop sample $(x = (s_{0:t}, c_0, n), y)$ from the dataset $\mathcal{D}$~(Eq.~\ref{eq:open-loop_samples}), the behavior model $\pi_\theta$ is unrolled to transform the ground truth input states $s_{0:t}$ into the closed-loop input states $s^{\text{cl}}_{0:t}$, thereby obtaining a closed-loop sample that incorporates the predicted behaviors of $\pi_\theta$:
}
\begin{equation}
(x^{\text{cl}} := (s^{\text{cl}}_{0:t}, c_0, n), y) \sim \mathcal{D}^{\text{cl}}.
\end{equation}

\replaced{
To ensure that the ground truth output $y$ serves as meaningful supervision, the closed-loop input $x^{\text{cl}}$ needs to stay close to the ground truth input $x$.
Therefore, at each time $h$ during closed-loop sample generation, based on the previously generated closed-loop states $x_{:h}^{\text{cl}} = (s^{\text{cl}}_{0:h}, c_0, n)$, we predict the future states $\hat{s}^{n}_{h:h+T_{\text{pred}}}$ using the following \textbf{closed-loop data generation policy}:
\begin{equation}
\pi_\text{data}^{\text{cl}}(\hat{s}^{n}_{h:h+T_{\text{pred}}} | x_{:h}^{\text{cl}}) := m_{\theta}(\hat{s}^{n}_{h:h+T_{\text{pred}}}| z_h^\text{cl}, x_{:h}^{\text{cl}}),
\end{equation}
which uses the prediction associated with component $z_h^{\text{cl}}$ in the mixture model $\pi_\theta$.
Here $z_h^{\text{cl}}$ is the component that best matches the ground truth states $s_{h:h+T_{\text{cl}}}^{n}$ over a \textbf{closed-loop matching window} of length $T_{\text{cl}}$:
\begin{equation}
z_h^{\text{cl}}\!=\!\arg\min_{k}\!
\left\{\!
\begin{aligned}
&d_{T_{\text{cl}}}\big(\mu_k(x_{:h}^{\text{cl}};\theta), s_{h:h+T_{\text{cl}}}^{n}\big), \ \text{anchor-free} \\
&d_{T_{\text{cl}}}\big(A_k(x_{:h}^{\text{cl}}), s_{h:h+T_{\text{cl}}}^{n}\big), \ \text{anchor-based}
\end{aligned}
\right.
.
\label{eq:posterior_component}
\end{equation}
The subscript of $d_{T_{\text{cl}}}$ highlights that the distance is computed over the shared time interval $T_{\text{cl}}$, which may be shorter than $T_{\text{pred}}$ in subsequent discussions.
}{
The specific closed-loop sample generation process is illustrated in Fig.~\ref{fig:closed-loop_samples}.
Starting from the initial ground truth states $s^{\text{cl}}_{0} = s_{0}$, a \textbf{posterior policy} $\pi^{\text{post}}$ is autoregressively applied:
\begin{equation}
\pi^{\text{post}}(x_{:h}^{\text{cl}}, s_{h:h+T_{\text{post}}}^{n}; \theta) := \mu_{z^{\text{post}}}(x_{:h}^{\text{cl}};\theta),
\label{eq:posterior_policy}
\end{equation}
where the posterior plan $\mu_{z^{\text{post}}}(x_{:h}^{\text{cl}};\theta)$ for time $h$ is the predicted trajectory corresponding to the component $z^{\text{post}}$ of $\pi_\theta$, based on the previously generated closed-loop input states in $x_{:h}^{\text{cl}} = (s^{\text{cl}}_{0:h}, c_0, n)$.
Similar to the positive component matching in Section~\ref{Positive Component Matching}, the posterior component $z^{\text{post}}$ is the one that best matches the ground truth $s_{h:h+T_{\text{post}}}^{n}$ over the \textbf{posterior planning horizon} $T_{\text{post}}$:
\begin{equation}
z^{\text{post}}\!=\!\arg\min_{k}\!
\left\{\!
\begin{aligned}
&d_{T_{\text{post}}}\big(\mu_k(x_{:h}^{\text{cl}};\theta), s_{h:h+T_{\text{post}}}^{n}\big), \ \text{anchor-free} \\
&d_{T_{\text{post}}}\big(A_k(x_{:h}^{\text{cl}}), s_{h:h+T_{\text{post}}}^{n}\big), \ \text{anchor-based}
\end{aligned}
\right.
.
\label{eq:posterior_component}
\end{equation}
Here, the subscript of $d_{T_{\text{post}}}$ highlights the distance is computed over the shared time interval $T_{\text{post}}$, which may be shorter than $T_{\text{pred}}$ in subsequent discussions.
}

\replaced{
In practice, we execute the mean trajectory $\mu_{z_h^{\text{cl}}}(x_{:h}^{\text{cl}};\theta)$ of the component $z_h^{\text{cl}}$ for each agent to generate the subsequent closed-loop states, and the replanning frequency is aligned with the simulation update interval $\tau$:
\begin{equation}
\begin{aligned}
\hat{s}^{n}_{h:h+T_{\text{pred}}} &= \pi_\text{data}^{\text{cl}}(x_{:h}^{\text{cl}}) = \mu_{z_h^{\text{cl}}}(x_{:h}^{\text{cl}};\theta), \\
s^{\text{cl}}_{h:h+\tau} &= \{\hat{s}^{n}_{h:h+\tau}\}_{n=1}^{N}.
\end{aligned}
\label{eq:execute_posterior_policy}
\end{equation}
}{
The posterior plans of each agent are then executed to generate the subsequent closed-loop states, and the replanning frequency is aligned with the simulation update interval $\tau$:
\begin{equation}
\begin{aligned}
s^{\text{cl}}_{h:h+\tau} &= \{\hat{s}^{n}_{h:h+\tau}\}_{n=1}^{N}.
\end{aligned}
\label{eq:execute_posterior_policy}
\end{equation}
}

When the above closed-loop sample generation is applied to \replaced{discrete NTP}{GPT-like discrete} models, namely anchor-based mixture models without continuous regression, we can discover that\deleted{:}
\textit{\replaced{in discrete NTP}{In GPT-like discrete} models~\cite{philion2024trajeglish,wu2024smart}, agent motion tokenization based on rolling matching is equivalent to closed-loop sample generation for\deleted{ discrete} mixture models}~(\textbf{H1}).
Therefore, \textit{most \replaced{discrete NTP models}{GPT-like models with categorical distributions} actually train on the closed-loop samples naturally introduced by tokenization}, as shown in Table~\ref{table:survey}.
In contrast, for mixture models with continuous regression~\cite{wang2023multiverse,zhou2024behaviorgpt}, although some works mention closed-loop training, they mostly refer to using an RNN-based motion decoder, while the model is in fact trained on open-loop samples.
Hence, the use of closed-loop samples is a commonly overlooked disparity between existing \replaced{regression-based mixture models and discrete NTP models}{continuous and discrete mixture models}, which could serve as a significant factor in explaining the performance gap.

\subsection{Shortcut Learning Issue} \label{Shortcut Learning Issue}
\replaced{For the closed-loop matching window $T_\text{cl}$, a natural choice is to set it equal to the prediction horizon $T_{\text{pred}}$, so that the selection of component $z^\text{cl}$ is performed over the same temporal span as the model’s prediction.}{Considering the goal of making input states more consistent between training and closed-loop simulation, the posterior planning horizon $T_{\text{post}}$ is by default set equal to the prediction horizon $T_{\text{pred}}$.}
However, when $T_{\text{pred}}$ exceeds the update interval $\tau$, meaning that the \replaced{matching window of $\pi_\text{data}^\text{cl}$}{planning horizon of $\pi^{\text{post}}$} is \replaced{longer}{greater} than its replanning interval~(\replaced{$T_{\text{cl}} = T_{\text{pred}} > \tau$}{$T_{\text{post}} > \tau$}), the generated closed-loop input $x^{\text{cl}}$ will incorporate information from the ground truth output $y$.
As shown in Fig.~\ref{fig:closed-loop_samples}(c), the closed-loop states $s^{\text{cl}}_{2\tau:3\tau}$ are derived based on \replaced{$s_{2\tau:2\tau+T_{\text{cl}}}$}{$s_{2\tau:2\tau+T_{\text{post}}}$} that overlap with the ground truth output.
This may lead the model $\pi_\theta$ to learn a shortcut, which could hinder its ability to generate realistic behaviors in\deleted{ closed-loop} simulation.

In the experiments, we attempt to set the \replaced{closed-loop matching window}{posterior planning horizon} equal to the update interval~(\replaced{$T_{\text{cl}} = \tau$}{$T_{\text{post}} = \tau$}) for resolving the potential shortcut learning issue.

\subsection{Off-Policy Learning Problem} \label{Off-Policy Learning Problem}
\deleted{
If one seeks to leverage a longer prediction horizon~($T_{\text{pred}} > \tau$), the aforementioned alignment between the posterior planning horizon and the update interval~($T_{\text{post}} = \tau$) will introduce a misalignment between $T_{\text{post}}$ and $T_{\text{pred}}$.
Upon closer inspection, this misalignment is primarily reflected in the fact that the behavior model $\pi_\theta$ is trained to select the positive component $z^*$ over $T_{\text{pred}}$~(Eq.~\ref{eq:anchor-free} and Eq.~\ref{eq:anchor-based}), while the posterior policy $\pi^\text{post}$ selects the posterior component $z^{\text{post}}$ over $T_{\text{post}}$~(Eq.~\ref{eq:posterior_component}).
This is analogous to the \textit{off-policy problem} in Reinforcement Learning~(RL), where a mismatch between the data collection policy and the training policy leads to distributional shifts.
Here, the off-policyness is mainly manifested in the disparity between the component selection horizons of $\pi_\theta$ and $\pi^\text{post}$.
}
\replaced{
To facilitate the subsequent discussion, we first introduce the \textbf{positive matching horizon} $T_{z^*}$ by extending Eq.~\ref{eq:anchor-free} and Eq.~\ref{eq:anchor-based} as follows:
}{
In fact, the component selection horizon of $\pi_\theta$, namely the \textbf{positive matching horizon} $T_{z^*}$, can differ from the prediction horizon $T_{\text{pred}}$ by extending Eq.~\ref{eq:anchor-free} and Eq.~\ref{eq:anchor-based} as follows:
}
\begin{equation}
z^*\!=\!\arg\min_{k}\!
\left\{\!
\begin{aligned}
&d_{T_{z^*}}\big(\mu_k(x^{\text{cl}};\theta), y\big), \ \text{anchor-free} \\
&d_{T_{z^*}}\big(A_k(x^{\text{cl}}), y\big), \ \text{anchor-based}
\end{aligned}
\right.
.
\label{eq:positive_matching_horizon}
\end{equation}
Similar to Eq.~\ref{eq:posterior_component}, $d_{T_{z^*}}$ represents the distance calculated over the first $T_{z^*}$ time interval of the trajectories.
This indicates that\replaced{, while $\pi_\theta$ generates trajectories over the prediction horizon $T_{\text{pred}}$, the positive component $z^*$ can be selected over a potentially shorter horizon $T_{z^*}$.}{ the positive component $z^*$ is selected over the horizon $T_{z^*}$, while $\pi_\theta$ could still generate trajectories over a longer $T_{\text{pred}}$.}

\added{
Traditionally, as indicated by Eq.~\ref{eq:anchor-free} and Eq.~\ref{eq:anchor-based}, the positive matching horizon coincides with the prediction horizon~($T_{z^*} = T_{\text{pred}}$).
To address the aforementioned shortcut learning issue, one may set the closed-loop matching window equal to the update interval~($T_{\text{cl}} = \tau$).
In this case, however, using a longer prediction horizon~($T_{\text{pred}} > \tau$) leads to a misalignment between $T_{\text{cl}}$ and $T_{z^*}$, since $T_{z^*}$ remains tied to $T_{\text{pred}}$ by default~($T_{z^*} = T_{\text{pred}} > T_{\text{cl}} = \tau$).
Upon closer inspection, this misalignment is primarily reflected in the fact that the model $\pi_\theta$ is trained to select the positive component $z^*$ over $T_{z^*}$~(Eq.~\ref{eq:positive_matching_horizon}), while the data generation policy $\pi_\text{data}^\text{cl}$ selects the component $z^{\text{cl}}$ over $T_{\text{cl}}$~(Eq.~\ref{eq:posterior_component}).
This is analogous to the \textit{off-policy} problem in Reinforcement Learning~(RL), where a mismatch between the data collection policy and the training policy leads to distributional shifts.
Here, the off-policyness is mainly manifested in the disparity between the component selection horizons of $\pi_\theta$ and $\pi_\text{data}^\text{cl}$.
}

\replaced{
Motivated by the above analysis, we decouple the positive matching horizon $T_{z^*}$ from the prediction horizon $T_\text{pred}$, and leverage the alignment between $T_{z^*}$ and $T_{\text{cl}}$ to study the off-policy learning problem in our experiments.
}{
In our experiments, we utilize the alignment between $T_{z^*}$ and $T_{\text{post}}$ to study the off-policy learning problem.
}

\subsection{\added{Disentanglement and Alignment of Horizons}} \label{Disentanglement and Alignment of Horizons}
\added{
Under the closed-loop data configuration, and building on the above discussion, we present the final training objective as formulated below:
\begin{equation}
\max_{\theta} \underbrace{\mathbb{E}_{(x^\text{cl},y) \sim \mathcal{D}^\text{cl}}}_{\pi_\text{data}^\text{cl}: \ T_\text{cl}=\tau} \Big[\underbrace{\log m_\theta(y|z^*,x^\text{cl})}_{\text{regression over } T_\text{pred}} - \underbrace{\mathbb{KL}[\hat{q}^*(z) \| q_\theta(z|x^\text{cl})]}_{\text{classification over } T_{z^*}}\Big].
\label{eq:final_objective}
\end{equation}
Following Eq.~\ref{eq:final_objective}, at each optimization step during training, we first generate a batch of closed-loop samples by unrolling $\pi_\text{data}^\text{cl}$, which selects the mixture component over $T_\text{cl}$ and replans every $\tau$.
We then update the model $\pi_\theta$ using these samples, so that the input states observed during training more closely resemble those encountered by the model in closed-loop simulation.
When computing the loss, $\pi_\theta$ selects the positive component $z^*$ over $T_{z^*}$, while the predicted trajectory associated with $z^*$ is generated and supervised over $T_{\text{pred}}$.
}

\added{
To effectively incorporate closed-loop samples with long prediction horizons---which constitutes the primary challenge in generalizing the use of closed-loop data---we disentangle and appropriately align the horizons used in data generation and model training:
}
\begin{itemize}[leftmargin=1em]
\item \added{For the shortcut learning issue, we decouple the closed-loop matching window $T_\text{cl}$ from the prediction horizon $T_{\text{pred}}$, and align $T_\text{cl}$ with the replanning interval $\tau$ of $\pi_\text{data}^\text{cl}$, yielding $T_{\text{pred}} > T_\text{cl}$ = $\tau$.}
\item \added{For the off-policy learning problem, we disentangle the positive matching horizon $T_{z^*}$ from the prediction horizon $T_{\text{pred}}$, and align $T_{z^*}$ with the closed-loop matching window $T_\text{cl}$, yielding $T_{\text{pred}} > T_{z^*} = T_\text{cl}$.}
\end{itemize}

\subsection{Approximate \replaced{Closed-Loop Data Generation}{Posterior} Policy} \label{Approximate Posterior Policy}
Except for discrete models, closed-loop sample generation involves iterative inference with the model $\pi_\theta$, which could take a relatively long time.
According to previous work~\cite{lin2024eda}, the behavior patterns generated by anchor-based models can be reflected in their anchors.
Therefore, we propose an \textbf{approximate \replaced{closed-loop data generation}{posterior} policy} for anchor-based models:
\replaced{
\begin{equation}
\pi_\text{data}^{\text{cl}}(x_{:h}^{\text{cl}}) \approx A_{z_h^{\text{cl}}}(x_{:h}^{\text{cl}}),
\end{equation}
where the anchor $A_{z_h^{\text{cl}}}(x_{:h}^{\text{cl}})$ rather than the predicted trajectory of the component $z_h^{\text{cl}}$ is applied.
}{
\begin{equation}
\pi^{\text{post}}(x_{:h}^{\text{cl}}, s_{h:h+T_{\text{post}}}^{n}; \theta) \approx A_{z^{\text{post}}}(x_{:h}^{\text{cl}}),
\end{equation}
where the anchor $A_{z^{\text{post}}}(x_{:h}^{\text{cl}})$ rather than the predicted trajectory of the component $z^{\text{post}}$ is applied.
}
In this way, the generation of closed-loop samples involves only predefined anchors, eliminating the need for $\pi_\theta$ computation and significantly reducing the required time.

In the experiments, we validate the efficiency and effectiveness of the above approximate \replaced{closed-loop data generation}{posterior} policy proposed for anchor-based mixture models.

\section{Model Architecture} \label{Model Architecture}

\begin{figure*}[tb]
\centering
\includegraphics[width=0.99\textwidth]{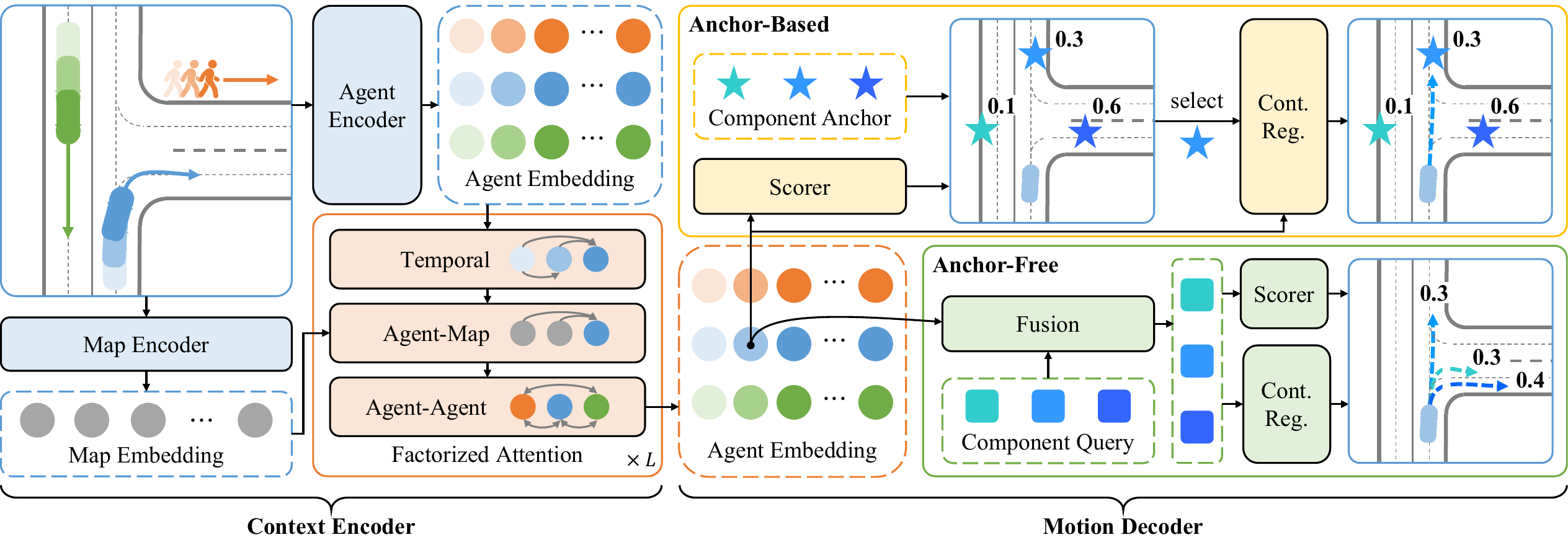}
\caption{
The demonstration of the model architecture, including the \textit{context encoder} and \textit{motion decoder}.
The motion decoders of anchor-based and anchor-free models are designed separately.
``Cont. Reg.'' represents the network for continuous regression, which outputs the trajectory of continuous states for the corresponding mixture component.
}
\label{fig:model_architecture}
\end{figure*}

The model architecture employed in our experiments, as demonstrated in Fig.~\ref{fig:model_architecture}, consists of a context encoder and a motion decoder.
The context encoder~(Section~\ref{Context Encoder}) can process information from multiple agents across multiple time steps in parallel.
The encoded embedding of the $n$-th agent at time $t$ can be viewed as the extracted features from the corresponding scene context $(S_{0:t}, C_0, n)$.
This agent embedding is then passed into the motion decoder to generate multimodal future motions for the $n$-th agent, starting from time $t$, along with the confidence scores for each component.
The motion decoders~(Section~\ref{Motion Decoder}) for the anchor-free and anchor-based models are separately designed.
Particularly, for anchor-based models with continuous regression, our decoder design enables scaling up the number of components as efficiently as in those without continuous regression.

\subsection{Context Encoder} \label{Context Encoder}
To efficiently process information from multiple agents across multiple time steps simultaneously, we adopt a symmetric scene context encoding based on query-centric attention~\cite{zhou2023query,shi2024mtr++}.
Specifically, for each scene element, such as a map polyline or an agent tracklet, the embedding is derived in its local reference frame.
When modeling interactions between scene elements, their relative positional encodings are integrated into the corresponding attention operations.
Following existing works~\cite{zhou2024behaviorgpt,wu2024smart}, we apply map self-attention within the map encoder, as well as the factorized attention containing temporal, agent-map and agent-agent attention~(Fig.~\ref{fig:model_architecture}), to obtain agent embeddings enriched with diverse spatio-temporal features.
During closed-loop simulation, thanks to the symmetric encoding~\cite{zhou2023query}, we can reuse previously derived embeddings to incrementally encode newly generated agent motions for faster inference, akin to the KV cache in LLMs~\cite{touvron2023llama}.

\subsection{Motion Decoder} \label{Motion Decoder}
Since the agent embeddings derived from the symmetric encoder contain spatio-temporal information in their local reference frame, the motion decoder treats each embedding equivalently and outputs trajectories in the corresponding local coordinate system.
Without loss of generality, focusing on the processing of an individual agent embedding, we next introduce the decoder designs for anchor-free and anchor-based models respectively.

\subsubsection{Anchor-Free Model}
For anchor-free models, similar to previous methods~\cite{zhou2024behaviorgpt}, each learnable query is linked to a specific component.
Every component query is fused with the agent embedding, which is then used to generate the corresponding trajectory along with its confidence score, as depicted in Fig.~\ref{fig:model_architecture}.
In the above process, the computational overhead of each part increases at least linearly with the number of components.
Besides, most existing anchor-based models with continuous regression~\cite{shi2024mtr++} adopt a decoder structure akin to anchor-free models, with the main difference being that component queries are generated using anchors.
It thus seems challenging for \replaced{regression-based}{general continuous} mixture models to scale up the number of components, as noted by methods employing discrete distributions~\cite{wu2024smart}.
In fact, for anchor-based models, the above situation has the potential to change, as will be discussed next.

\subsubsection{Anchor-Based Model}
For anchor-based models, we first generate anchor trajectories for each agent category by clustering the training data\deleted{~(Fig.~\ref{fig:kmeans_anchors})}, as done in previous works~\cite{wang2023multiverse,philion2024trajeglish}.
In this context, for a specific agent category, the anchor corresponding to each component index is actually well-defined.
Additionally, the confidence scores in anchor-based models assess the congruence between the ground truth and anchors, rather than predicted trajectories.
Therefore, we can directly use the agent embedding to predict a categorical distribution that represents the scores for each anchor, as shown in Fig.~\ref{fig:model_architecture}.
When employing continuous regression, we only need to select a single anchor, either the one associated with the positive component or one sampled based on the scores, and generate its corresponding trajectory.
If continuous regression is not applied, the model is equivalent to one that uses a discrete distribution~\cite{wu2024smart}.
While preserving the features inherent to anchor-based models, the above design markedly mitigates the increase in decoder computational cost as the number of components grows.
Such a design also unifies continuous and discrete anchor-based models, thereby facilitating the investigation of continuous regression.


\section{Experiments}
In the experiments, we begin by systematically examining the effects of varying prediction horizons~(Section~\ref{Exp: Prediction Horizon}) and component numbers~(Section~\ref{Exp: Number of Components}) for both anchor-free and anchor-based models.
The models in the above experiments~(Section~\ref{Training with Open-Loop Samples}) are all trained on the original open-loop samples\deleted{ to ensure consistency in the training data}.
Next, we select the optimal prediction horizons and component numbers\deleted{ for the anchor-free and anchor-based models respectively,} to explore training with closed-loop samples~(Section~\ref{Training with Closed-Loop Samples}).
Specifically, we investigate the shortcut learning issue~(Section~\ref{Exp: Shortcut Learning Issue}) and the off-policy learning problem~(Section~\ref{Exp: Off-Policy Learning Problem}).
Additionally, for anchor-based models, we \replaced{examine}{validate} the effectiveness of the proposed approximate \replaced{closed-loop data generation}{posterior} policy~(Section~\ref{Exp: Approximate Posterior Policy}) and\deleted{ examine the performance gain introduced by} continuous regression~(Section~\ref{Exp: Continuous Regression}).
\added{We also quantify the efficiency impact of training with closed-loop samples~(Section~\ref{Training-Time Overhead}).}
From the above exploratory experiments, we \replaced{summarize the key designs for realistic simulation~(Section~\ref{Key Designs for Realistic Simulation}) and evaluate distinct UniMM variants }{select the best configurations that include the anchor-free model as well as anchor-based models with and without continuous regression, evaluating their performance }on the WOSAC~\cite{montali2024waymo} benchmark~(Section~\ref{Benchmark Results}).
\added{
In addition, we provide qualitative results~(Section~\ref{Qualitative Results}), and further integrate UniMM into the MetaDrive~\cite{li2022metadrive} simulator for simulation-based evaluation~(Section~\ref{Toward Real-World System Application}).
}

\subsection{Experimental Setup}
\subsubsection{Dataset}
We perform experiments on the large-scale Waymo Open Motion Dataset~(WOMD)~\cite{ettinger2021large}, which includes 487k training, 44k validation, and 44k testing scenes.
Each scene contains an HD map and up to 128 agents, spans 9 seconds, and is sampled at a frequency of 10 Hz, with 1 second of historical data and 8 seconds of future trajectories.
For efficiently conducting the exploratory experiments in Section~\ref{Training with Open-Loop Samples} and Section~\ref{Training with Closed-Loop Samples}, we train models on 20\% of the scenes uniformly sampled from the training set, which is empirically found to exhibit a similar distribution to the full training set~\cite{shi2024mtr++}.
In Section~\ref{Benchmark Results}, the models evaluated on the benchmark are trained on the full WOMD training set.

\subsubsection{Metrics}
For \textit{multi-agent simulation}, we adopt the evaluation metrics from the Waymo Open Sim Agents Challenge~(WOSAC)~\cite{montali2024waymo}, which include \replaced{likelihoods over kinematic, interactive, and map-based features}{Kinematic metrics~(likelihood of linear speed and acceleration, angular speed and acceleration), Interactive metrics~(likelihood of distance to nearest object, collision, time to collision), and Map-based metrics~(likelihood of distance to road edge, offroad)}.
The Realism Meta metric serves as an overall measure, derived from a weighted combination of the above metrics.
The WOSAC metrics are by default based on 32 rollouts, each lasting 8 seconds at 10 Hz, for every scenario with 1 second of history from the WOMD validation or testing set.
As the number of scenes is large and the computation of metrics is expensive, evaluating on the full set\deleted{ via the official website} is time-consuming.
For the exploratory experiments in Section~\ref{Training with Open-Loop Samples} and Section~\ref{Training with Closed-Loop Samples}, we evaluate models on \replaced{a validation subset that contains 150 scenes in total}{a subset of the validation set, using a local evaluation tool based on the official API to compute the metrics.
The subset consists of the first scene from each shard of the dataset, containing 150 scenes in total}.
The benchmark results in Section~\ref{Benchmark Results} are statistics on the full validation and testing sets.
In some cases, we also assess the \textit{open-loop prediction} performance as a reference, using metrics from the Waymo Motion Prediction Challenge~\cite{ettinger2021large}, including \replaced{minimum errors, miss rate, and mAP}{minADE~(Minimum Average Displacement Error), minFDE~(Minimum Final Displacement Error), MR~(Miss Rate), and mAP~(Mean Average Precision)}.
\added{Further details on the evaluation metrics can be found in Appendix~\replaced{A}{\ref*{appendix:metrics}}.}
\deleted{
The motion prediction metrics are computed on multiple 8-second future trajectories of interested agents, given 1 second of history.
We apply the official evaluation tool locally to calculate the metrics on the full WOMD validation set, since the evaluation server only accepts submissions with up to 6 trajectories for each target agent.
}

\subsubsection{Implementation Details}
\deleted{
To balance granularity and efficiency, we downsample the HD map so that the point distance is around 2.5 meters and divide it into polylines with intervals of about 5 meters.
The agent trajectories are segmented into multiple tracklets, each with a duration equal to the simulation update interval $\tau = 0.5s$.
The embeddings at the map polyline or agent tracklet level are derived through attention-based aggregation.
}
For modeling interactions among scene elements, we apply 1 layer of map self-attention and stack 2 layers of factorized attention.
The scorer and continuous regression networks in the decoder are implemented using 2-layer MLPs.
In anchor-free models, component queries and agent embeddings are fused by concatenation.
In anchor-based models, anchors are generated through the widely-used k-means clustering on the training set~\cite{chai2019multipath}\deleted{, as shown in Figure~\ref{fig:kmeans_anchors}}.
Before entering the continuous regression network, the selected anchor is encoded by a 2-layer MLP and then concatenated with the agent embedding.
Ultimately, the models in our experiments generally have around 4 million parameters and comparable computational overhead~(Table~\replaced{X}{\ref*{table:consumption}} \added{in Appendix~\replaced{C-B}{\ref*{appendix:consumption}}}).
\added{More implementation details can be found in Appendix~\replaced{B}{\ref*{appendix:details}}.}
\deleted{
The agent state in the model's predicted trajectory includes 2D position and heading.
Following previous works~\cite{shi2022motion,zhou2023query}, we treat each coordinate and time step in the trajectory as independent.
In the component distribution, we use Laplace distributions for position and von Mises distribution for heading.
At inference, the output trajectory for a specific component utilizes the expected value of its corresponding component distribution.
During closed-loop simulation, we sample the components based on the original output probabilities to avoid introducing bias and ensure a fair comparison.
}

\subsubsection{Training Details}
The models are trained end-to-end using the AdamW optimizer with a weight decay of 0.0001.
Training is performed with a batch size of 32 scenes for 30 epochs on 8 GPUs~(NVIDIA RTX 4090).
The learning rate is decayed from 0.0005 to 0 using a cosine annealing scheduler.
Consistent with simulation, the scene's $1s$ history is included in the initial context $C_0$, and all valid agents at the current time are used.
Historical states in $C_0$ are processed by the encoder like other states, except that their agent embeddings are not passed to the decoder for prediction.
Using the $8s$ future from the training data, the model predicts in parallel based on corresponding agent embeddings at intervals of $\tau = 0.5s$.

\subsection{Training with Open-Loop Samples} \label{Training with Open-Loop Samples}
In this section, we evaluate anchor-free models with 3, 6, and 16 components, as well as anchor-based models with 64, 512, and 2048 components.
At the same time, we investigate prediction horizons of 0.5s, 2s, 4s, and 8s for the aforementioned models.
All of these models are trained on consistent open-loop data samples, thereby explicitly reflecting the characteristics of the configurations under investigation.
The corresponding results are shown in Table~\ref{table:open-loop}.

\begin{table*}[tb]
\centering
\caption{
WOSAC Metrics of Models with Varying Prediction Horizons and Component Numbers on a Subset of WOMD Validation Set
}
\resizebox{0.88\textwidth}{!}{
\def\arraystretch{1.1}
\begin{threeparttable}
\begin{tabular}{c|c|c|ccccc}
\specialrule{1pt}{0pt}{0pt}
\multirow{2}{*}{\makecell{Anchor-\\Based}} & \multirow{2}{*}{\makecell{Number of\\Components}} & \multirow{2}{*}{\makecell{Prediction\\Horizon}} & \multirow{2}{*}{\textbf{Realism Meta} $\uparrow$} & \multirow{2}{*}{Kinematic $\uparrow$} & \multirow{2}{*}{Interactive $\uparrow$} & \multirow{2}{*}{Map-based $\uparrow$} & \multirow{2}{*}{minADE $\downarrow$} \\ & & & & & & & \\
\hline
\multirow{12}{*}{$\times$} 
& \multirow{4}{*}{3} 
& 0.5s & 0.7071 & 0.3942 & 0.7756 & 0.7977 & 1.8274 \\
& & 2s & \textbf{0.7388} & \textbf{0.4099} & 0.7966 & \textbf{0.8524} & 1.6001 \\
& & 4s & 0.7382 & 0.4096 & \textbf{0.8014} & 0.8446 & \textbf{1.5005} \\
& & 8s & 0.7355 & 0.4043 & 0.7974 & 0.8452 & 1.5960 \\
\cline{2-8}
& \multirow{4}{*}{6} 
& 0.5s & 0.6948 & 0.4010 & 0.7529 & 0.7880 & 2.2808 \\ 
& & 2s & 0.7385 & \textbf{0.4266} & 0.7914 & \textbf{0.8488} & 1.6000 \\
& & 4s & \textbf{0.7409} & 0.4212 & \textbf{0.8047} & 0.8416 & 1.5971 \\
& & 8s & 0.7347 & 0.4185 & 0.7933 & 0.8401 & \textbf{1.5960} \\
\cline{2-8}
& \multirow{4}{*}{16}
& 0.5s & 0.6269 & 0.4126 & 0.6640 & 0.7018 & 3.2278 \\ 
& & 2s & 0.7266 & 0.4345 & 0.7685 & 0.8395 & 1.9114 \\
& & 4s & 0.7343 & \textbf{0.4348} & 0.7811 & \textbf{0.8453} & 1.7381 \\
& & 8s & \textbf{0.7361} & 0.4286 & \textbf{0.7899} & 0.8428 & \textbf{1.6997} \\
\hline
\multirow{12}{*}{\checkmark} 
& \multirow{4}{*}{64} 
& 0.5s & 0.7097 & 0.3744 & 0.7831 & 0.8070 & 2.1082 \\ 
& & 2s & \textbf{0.7431} & \textbf{0.4203} & 0.7977 & \textbf{0.8573} & 1.5491 \\
& & 4s & 0.7388 & 0.4168 & \textbf{0.8049} & 0.8378 & \textbf{1.5077} \\
& & 8s & 0.7364 & 0.4147 & 0.7905 & 0.8506 & 1.5618 \\
\cline{2-8}
& \multirow{4}{*}{512} 
& 0.5s & 0.7318 & 0.4217 & 0.7902 & 0.8340 & 1.8040 \\ 
& & 2s & \textbf{0.7512} & \textbf{0.4402} & \textbf{0.8042} & \textbf{0.8608} & \textbf{1.5122} \\
& & 4s & 0.7481 & 0.4369 & 0.8037 & 0.8544 & 1.5384 \\
& & 8s & 0.7460 & 0.4350 & 0.8015 & 0.8523 & 1.5927 \\
\cline{2-8}
& \multirow{4}{*}{2048} 
& 0.5s & 0.7361 & 0.4354 & 0.7902 & 0.8384 & 1.8024 \\ 
& & 2s & 0.7537 & 0.4542 & \textbf{0.8086} & 0.8543 & 1.4646 \\
& & 4s & \textbf{0.7570} & \textbf{0.4546} & 0.8072 & \textbf{0.8653} & \textbf{1.4187} \\
& & 8s & 0.7489 & 0.4519 & 0.7992 & 0.8540 & 1.4939 \\
\specialrule{1pt}{0pt}{0pt}
\end{tabular}
\begin{tablenotes}[flushleft]
\item The models are trained on open-loop data samples.
\end{tablenotes}
\end{threeparttable}
}
\label{table:open-loop}
\vspace{-11pt}
\end{table*}

\subsubsection{Prediction Horizon} \label{Exp: Prediction Horizon}

\begin{figure}[tb]
\centering
\includegraphics[width=0.99\columnwidth]{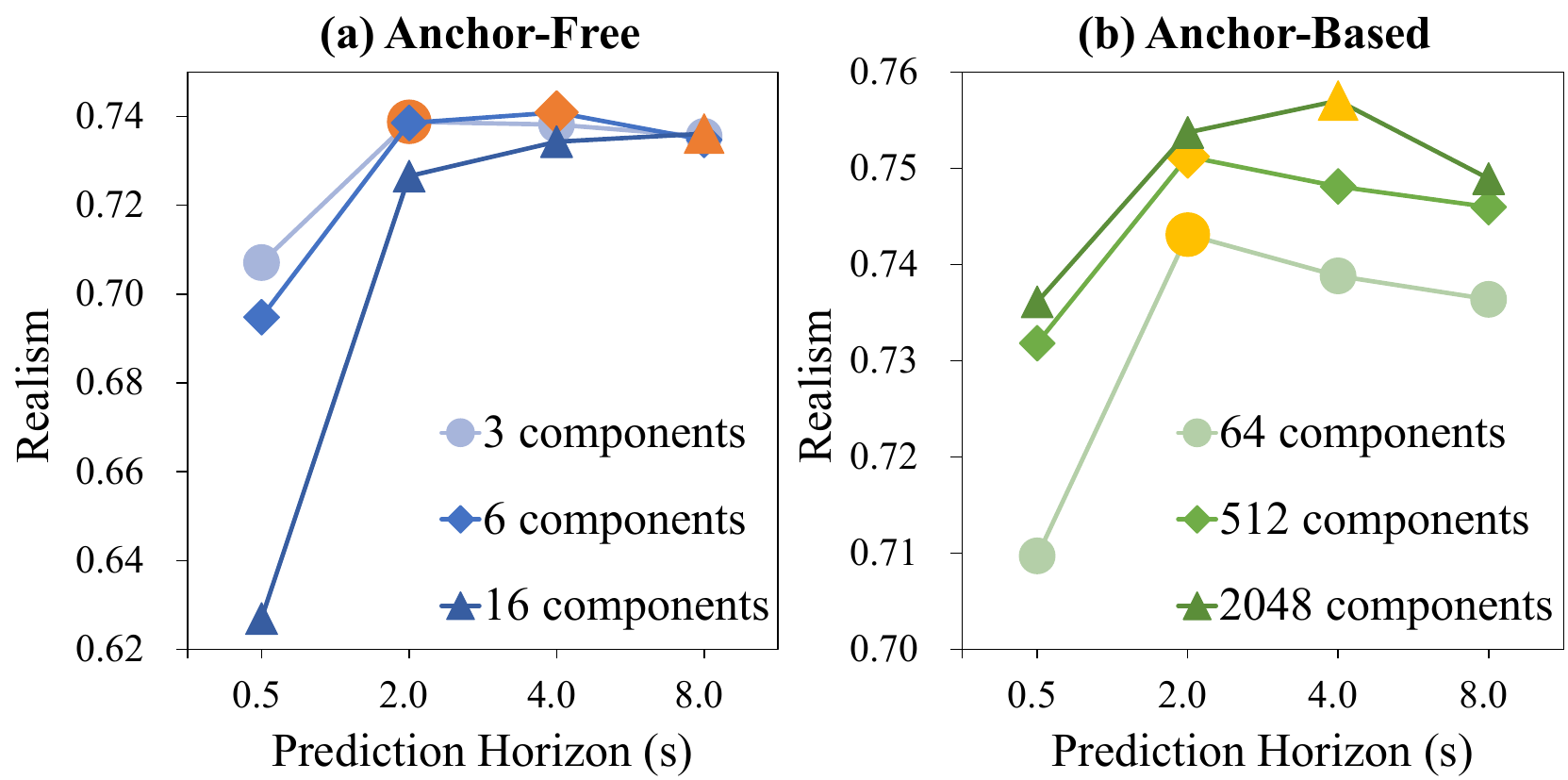}
\caption{
Realism Meta metric for models with different prediction horizons.
The models are trained on open-loop data samples.
}
\label{fig:prediction_horizon}
\vspace{-11pt}
\end{figure}

When the prediction horizon exceeds $\tau = 0.5s$, predictions starting from later time steps may not have sufficiently long ground truth trajectories for matching and supervision.
In such cases, we use the valid ground truth states within the corresponding prediction time window.
For further alignment, the anchor-based models with the same number of components and different prediction horizons are based on the same set of $8s$ anchors.
Specifically, the model uses the initial portion of the $8s$ anchor trajectory corresponding to the prediction horizon as its anchor.

As shown in Table~\ref{table:open-loop} and Fig.~\ref{fig:prediction_horizon}, a longer prediction horizon~(from 0.5s to 2s) initially leads to significant improvements for all models, which is also observed in previous works~\cite{suo2021trafficsim, hu2025solving}.
However, when the prediction horizon is further extended, the improvements in metrics become less pronounced and may even decline.
The above results suggest \added{two complementary findings}:
\added{(1)}~\textit{The additional supervision signal provided by increasing the prediction horizon is effective}\replaced{;}{.}
\added{(2)}~\textit{Nevertheless, an excessively large prediction horizon may lead the model to focus more on optimizing distant predictions, which could be detrimental as only the initial segment of the predicted trajectory is utilized during simulation}.
Additionally, we find that the optimal prediction horizon for simulation realism increases with the number of components, both for anchor-free and anchor-based models, which is consistent with the positive correlation between future time length and uncertainty.

\subsubsection{Number of Components} \label{Exp: Number of Components}
\begin{figure}[tb]
\centering
\includegraphics[width=0.99\columnwidth]{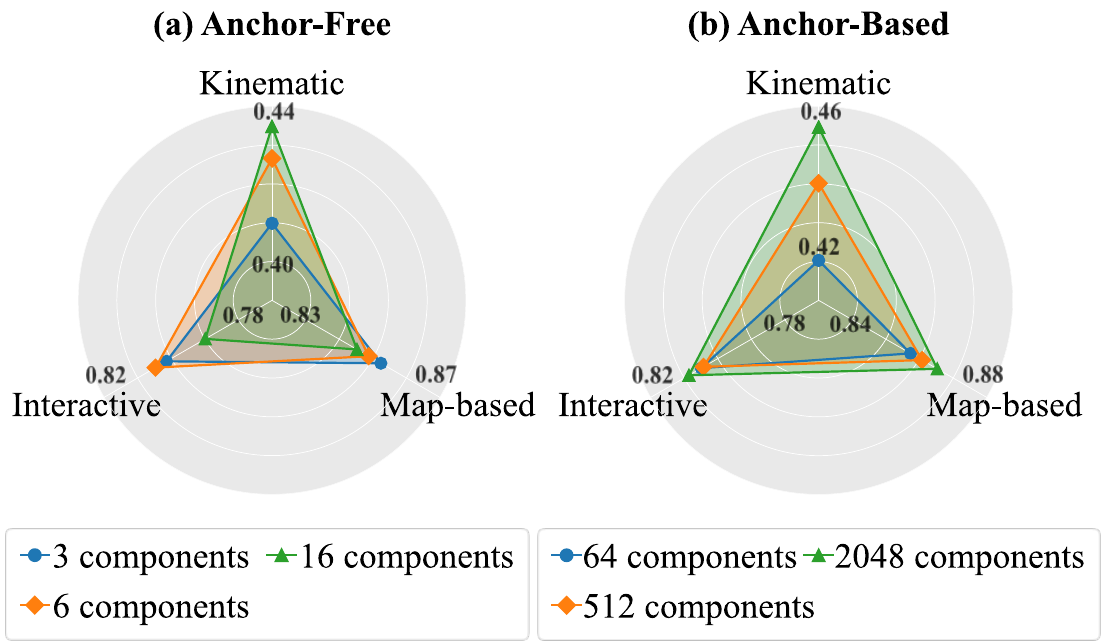}
\caption{
The optimal values of each WOSAC metric across prediction horizons for models with varying component numbers.
The models are trained on open-loop data samples.
}
\label{fig:component_number}
\vspace{-11pt}
\end{figure}

To more clearly demonstrate the impact of the number of components, we plot the optimal metrics across prediction horizons~(bold numbers in Table~\ref{table:open-loop}) for models with different component numbers in Fig.~\ref{fig:component_number}.

\textit{For anchor-free models,} as shown in Table~\ref{table:open-loop}, increasing the number of components from 3 to 6 provides some improvements in simulation realism, but further increasing it to 16 could lead to a decline.
Based on previous works~\cite{varadarajan2022multipath++, nayakanti2023wayformer}, we suspect that it may be challenging for anchor-free models to select better components directly using the output scores, when the number of components is large.
We thus also evaluate the motion prediction metrics for anchor-free models\deleted{~(Table~\ref{table:anchor-free_pred})} to facilitate validation\added{, with additional results provided in Appendix~\replaced{C-A}{\ref*{appendix:motion_prediction}}}.
As the number of components increases, the kinematic metrics of the simulations improve~(Fig.~\ref{fig:component_number}(a)), and the minimum errors as well as the miss rate in the predictions decrease~(Table~\replaced{IX}{\ref*{table:anchor-free_pred}} \added{in Appendix~\replaced{C-A}{\ref*{appendix:motion_prediction}}}).
This suggests \replaced{that}{:}
\textit{\replaced{a}{A} higher number of components can indeed enhance the ability to represent complex distributions}, as implied by existing methods~\cite{zhou2024behaviorgpt}.
However, other metrics of the simulations do not continue to improve as the number of components increases~(Fig.~\ref{fig:component_number}(a)).
Meanwhile, the mAP\replaced{---reflecting }{, which reflects }the ability to score multimodal motions\replaced{---}{, }consistently deteriorates with the increase in component number\replaced{, and}{~(Table~\ref{table:anchor-free_pred}).
Moreover, as shown in Table~\ref{table:anchor-free_pred},} the top-6 predictions of the model with 16 components are even worse than the model with 3 components across all prediction metrics\added{~(Table~IX in Appendix~C-A)}.
The above results suggest that\deleted{:}
\textit{\replaced{a}{A} larger number of components may hinder anchor-free models from picking out realistic trajectories, thereby impacting their simulation performance}.

\textit{For anchor-based models,} according to existing studies~\cite{shi2024mtr++,lin2024eda}, anchors as motion priors considerably alleviate the challenge of scoring multimodal predictions.
Furthermore, since anchor-based models tend to generate predictions around the anchors, increasing anchor density helps improve trajectory accuracy.
As shown in Table~\ref{table:open-loop}\replaced{,}{:}
\textit{\replaced{anchor-based}{Anchor-based} models consistently benefit from the increase in the number of components, across various prediction horizons~(Fig.~\ref{fig:prediction_horizon}(b)) and simulation metrics~(Fig.~\ref{fig:component_number}(b))}.

Additionally, as indicated by Table~\replaced{X}{\ref*{table:consumption}} \added{in Appendix~\replaced{C-B}{\ref*{appendix:consumption}}}, the computational costs of anchor-free models increase steadily with the number of components, such that the model with 64 components encounters memory exhaustion during training.
In contrast, thanks to the architecture design in Section~\ref{Motion Decoder}, \textit{anchor-based models can easily scale up the number of components} without a notable increase in computational overhead.

\subsection{Training with Closed-Loop Samples} \label{Training with Closed-Loop Samples}
Among the models trained with open-loop data samples in Section~\ref{Training with Open-Loop Samples}, the \textit{anchor-free model with 6 components} and the \textit{anchor-based model with 2048 components}, both using a \textit{4s prediction horizon}, exhibit the best simulation realism in their respective positive component matching paradigms.
In this section, we adopt the above configurations as baselines to highlight the impact of closed-loop samples.
For a fair comparison, unless otherwise specified, models leveraging closed-loop samples are also trained from scratch.
To ensure sample quality, during each iteration of closed-loop sample generation~(Eq.~\ref{eq:execute_posterior_policy}), \replaced{the predicted trajectory of $\pi_\text{data}^\text{cl}$}{a posterior plan} is executed only if its error falls below a certain threshold; otherwise, the corresponding ground truth states are retained.

\begin{table*}[tb]
\setlength\tabcolsep{4pt}
\centering
\caption{
Shortcut Learning Issue and Off-Policy Learning Problem for Models Trained with Closed-Loop Samples
}
\resizebox{0.99\textwidth}{!}{
\def\arraystretch{1.2}
\begin{threeparttable}
\begin{tabular}{c|c|cccc|ccccc}
\specialrule{1pt}{0pt}{0pt}
\multirow{3}{*}{\makecell{Anchor-\\Based}} & \multirow{3}{*}{\makecell{Prediction\\Horizon}} & \multirow{3}{*}{\makecell{Closed-\\Loop\\Samples}} & \multirow{3}{*}{\makecell{\replaced{Closed-Loop}{Posterior}\\ \replaced{Matching}{Planning}\\ \replaced{Window}{Horizon}}} & \multirow{3}{*}{\makecell{Positive\\Matching\\Horizon}} & \multirow{3}{*}{\makecell{Additional\\Scorer\\$@\tau$}} & \multirow{3}{*}{\textbf{Realism Meta} $\uparrow$} & \multirow{3}{*}{Kinematic $\uparrow$} & \multirow{3}{*}{Interactive $\uparrow$} & \multirow{3}{*}{Map-based $\uparrow$} & \multirow{3}{*}{minADE $\downarrow$} \\ & & & & & & & & & & \\ & & & & & & & & & & \\
\hline
\multirow{7}{*}{$\times$} 
& \multirow{5}{*}{4s $> \tau$}
& $\times$ & - & 4s & - & 0.7409 & 0.4212 & 0.8047 & 0.8416 & 1.5971 \\
& & $\checkmark$ & 4s & 4s & - & 0.7376 & 0.4450 & 0.7753 & 0.8563 & 1.5970 \\
& & \gcell{$\checkmark$} & \gcell{0.5s} & \gcell{4s} & \gcell{$\times$}
& \gcell{\textbf{0.7624}} & \gcell{0.4718} & \gcell{\textbf{0.8096}} & \gcell{\textbf{0.8679}} & \gcell{\textbf{1.3967}} \\
& & \gcell{$\checkmark$} & \gcell{0.5s} & \gcell{4s} & \gcell{$\checkmark$}
& \gcell{\textbf{0.7623}} & \gcell{0.4758} & \gcell{\textbf{0.8093}} & \gcell{0.8655} & \gcell{1.4385} \\
& & \gcell{$\checkmark$} & \gcell{0.5s} & \gcell{0.5s} & \gcell{-}
& \gcell{0.7583} & \gcell{\textbf{0.4834}} & \gcell{0.8028} & \gcell{0.8582} & \gcell{1.5598} \\
\cline{2-11}
& \multirow{2}{*}{0.5s $= \tau$}
& $\times$ & - & 0.5s & - & 0.6948 & 0.4010 & 0.7529 & 0.7880 & 2.2808 \\
& & $\checkmark$ & 0.5s & 0.5s & - & 0.7404 & 0.4737 & 0.7790 & 0.8432 & 1.8247 \\
\hline
\multirow{7}{*}{$\checkmark$} 
& \multirow{5}{*}{4s $> \tau$}
& $\times$ & - & 4s & - & 0.7570 & 0.4546 & 0.8072 & 0.8653 & 1.4187 \\
& & $\checkmark$ & 4s & 4s & - & 0.7475 & 0.4774 & 0.7828 & 0.8566 & 1.5128 \\
& & \gcell{$\checkmark$} & \gcell{0.5s} & \gcell{4s} & \gcell{$\times$}
& \gcell{0.7504} & \gcell{0.4727} & \gcell{0.7947} & \gcell{0.8522} & \gcell{1.8839} \\
& & \gcell{$\checkmark$} & \gcell{0.5s} & \gcell{4s} & \gcell{$\checkmark$}
& \gcell{0.7577} & \gcell{0.4566} & \gcell\textbf{0.8086} & \gcell{0.8644} & \gcell{1.4339} \\
& & \gcell{$\checkmark$} & \gcell{0.5s} & \gcell{0.5s} & \gcell{-}
& \gcell{\textbf{0.7666}} & \gcell{\textbf{0.4997}} & \gcell{0.8059} & \gcell{\textbf{0.8686}} & \gcell{\textbf{1.3708}} \\
\cline{2-11}
& \multirow{2}{*}{0.5s $= \tau$}
& $\times$ & - & 0.5s & - & 0.7361 & 0.4354 & 0.7902 & 0.8384 & 1.8024 \\ 
& & $\checkmark$ & 0.5s & 0.5s & - & \textbf{0.7658} & \textbf{0.5005} & 0.8054 & 0.8665 & 1.4062 \\
\specialrule{1pt}{0pt}{0pt}
\end{tabular}
\begin{tablenotes}[flushleft]
\item
The anchor-free models are trained with 6 components, while the anchor-based models are trained with 2048 components.
The metrics are computed on the validation subset of 150 scenes.
The rows highlighted in gray correspond to experiments aimed at reducing off-policyness.
\end{tablenotes}
\end{threeparttable}
}
\label{table:closed-loop}
\vspace{-11pt}
\end{table*}

\subsubsection{Shortcut Learning Issue} \label{Exp: Shortcut Learning Issue}
As shown in Table~\ref{table:closed-loop}, when the \replaced{closed-loop matching window}{posterior planning horizon} equals the prediction horizon and exceeds the update interval~(\replaced{$T_{\text{cl}} = T_{\text{pred}} > \tau$}{$T_{\text{post}} = T_{\text{pred}} > \tau$}), both the anchor-free and anchor-based models trained on closed-loop samples exhibit reduced simulation realism.
Specifically, the kinematic metrics show an increase, while the interactive metrics decrease significantly.
On the other hand, if we set \replaced{$T_{\text{cl}} = T_{\text{pred}} = \tau$}{$T_{\text{post}} = T_{\text{pred}} = \tau$}, training with closed-loop samples results in a considerable improvement, with the anchor-based model reaching state-of-the-art performance.
Moreover, we attempt to align \replaced{$T_{\text{cl}} = \tau$}{$T_{\text{post}} = \tau$} while maintaining $T_{\text{pred}} > \tau$, and it turns out that the anchor-free model demonstrates a competitive outcome.
The above results indicate \replaced{the following}{that}:
\textit{When the \replaced{closed-loop data generation}{posterior} policy’s \replaced{matching window}{planning horizon} exceeds its replanning interval~(\replaced{$T_{\text{cl}} > \tau$}{$T_{\text{post}} > \tau$}), the model indeed learns a shortcut, which compromises its reasoning of spatio-temporal interactions}, as noted in Section~\ref{Shortcut Learning Issue}.

\subsubsection{Off-Policy Learning Problem} \label{Exp: Off-Policy Learning Problem}
When setting \replaced{$T_{\text{cl}} = \tau$}{$T_{\text{post}} = \tau$} to address the shortcut learning issue, as shown in Table~\ref{table:closed-loop}, a longer prediction horizon ($T_{\text{pred}} > \tau$) is beneficial for the anchor-free model, but not necessarily for the anchor-based model.
We suspect this is due to the \textit{off-policy problem} arising from the misalignment between the positive matching horizon $T_{z^*}$~(Eq.~\ref{eq:positive_matching_horizon}) and the \replaced{closed-loop matching window $T_{\text{cl}}$}{posterior planning horizon $T_{\text{post}}$}~(Eq.~\ref{eq:posterior_component}), as discussed in Section~\ref{Off-Policy Learning Problem}.
Therefore, based on the trained model with \replaced{$T_{\text{pred}} > T_{\text{cl}} = \tau$}{$T_{\text{pred}} > T_{\text{post}} = \tau$}, we \textbf{train an additional scorer to select $\boldsymbol{z^*}$ over $\boldsymbol{T_{z^*} = \tau}$}, while keeping the previously trained parts frozen.
Table~\ref{table:closed-loop} shows that using the additional scorer$@\tau$ instead of the original scorer$@T_{\text{pred}}$ improves the simulation realism of the anchor-based model, while the anchor-free model remains largely unchanged.
The above results imply \added{two key observations}:
\added{(1)}~\textit{\replaced{These results provide preliminary}{Preliminary} validation of the above speculation that the off-policy learning problem~(Section~\ref{Off-Policy Learning Problem}) could hinder closed-loop samples from benefiting models}\replaced{;}{.}
\added{(2)}~\textit{For anchor-free models, the off-policy problem appears less severe, probably because they rely more on flexible predictions for each component rather than component selection}.

Despite improvement in the meta metric, the anchor-based model with the additional scorer$@\tau$ shows reduced kinematic metrics, possibly due to the anchor selected over $T_{z^*} = \tau$ being unfamiliar for the pre-trained continuous regression network.
Hence, we attempt to \textbf{train the model from scratch with \replaced{$\boldsymbol{T_{\text{pred}} > T_{z^*} = T_{\text{cl}} = \tau}$}{$T_{\text{pred}} > T_{z^*} = T_{\text{post}} = \tau$}}, where the predicted trajectory for the positive component $z^*$, selected over $\tau$, is generated and supervised over $T_{\text{pred}}$.
Intuitively, the optimization difficulty of regression in Eq.~\ref{eq:training_objective} increases, since the trajectory of $z^*$ selected over a shorter horizon typically has a larger error.
The anchor-free model thus experiences a modest decline, as shown in Table~\ref{table:closed-loop}.
On the other hand, the anchor-based model with \replaced{$T_{\text{pred}} > T_{z^*} = T_{\text{cl}} = \tau$}{$T_{\text{pred}} > T_{z^*} = T_{\text{post}} = \tau$} performs even slightly better than the one with \replaced{$T_{\text{pred}} = T_{z^*} = T_{\text{cl}} = \tau$}{$T_{\text{pred}} = T_{z^*} = T_{\text{post}} = \tau$}.
The above evidence further demonstrates that\deleted{:}
\textit{\replaced{aligning}{Aligning} component selection horizons~(\replaced{$T_{z^*} = T_{\text{cl}}$}{$T_{z^*} = T_{\text{post}}$}) can effectively mitigate the off-policy problem, particularly for anchor-based models that heavily rely on anchor selection}.


\subsubsection{Approximate \replaced{Closed-Loop Data Generation}{Posterior} Policy} \label{Exp: Approximate Posterior Policy}
In Table~\ref{table:closed-loop}, the anchor-based models with 0.5s and 4s prediction horizons, both configured with \replaced{$T_{\text{cl}} = T_{z^*} = \tau$}{$T_{\text{post}} = T_{z^*} = \tau$}, achieve the top performance.
Therefore, we attempt to apply the approximate \replaced{closed-loop data generation}{posterior} policy~(Section~\ref{Approximate Posterior Policy}) to both models.
As shown in Table~\ref{table:approximate_posterior_policy}, 
\textit{\replaced{the}{The} anchor-based models utilizing the approximate \replaced{closed-loop data generation}{posterior} policy achieve comparable performance while requiring significantly less training time}.

\begin{table}[tb]
\setlength\tabcolsep{2pt}
\centering
\caption{
Approximate \replaced{Closed-Loop Data Generation}{Posterior} Policy for Anchor-Based Models
}
\resizebox{0.99\columnwidth}{!}{
\def\arraystretch{1.1}
\begin{threeparttable}
\begin{tabular}{cc|cccc|c}
\specialrule{1pt}{0pt}{0pt}
\multirow{3}{*}{\makecell{Prediction\\Horizon}} & \multirow{3}{*}{\makecell{Approx.\\ \replaced{$\pi_\text{data}^\text{cl}$}{Posterior\\Policy}}} & \multirow{3}{*}{\textbf{Meta}} & \multirow{3}{*}{Kinematic} & \multirow{3}{*}{Interactive} & \multirow{3}{*}{Map-based} & \multirow{3}{*}{\makecell{Training\\Time}} \\ & & & & & & \\ & & & & & & \\
\hline
\multirow{2}{*}{4s}
& $\times$ & 0.7666 & 0.4997 & 0.8059 & 0.8686 & $\sim 46h$ \\
& $\checkmark$ & 0.7686 & 0.4950 & 0.8138 & 0.8667 & $\sim 16h$ \\
\hline
\multirow{2}{*}{0.5s}
& $\times$ & 0.7658 & 0.5005 & 0.8054 & 0.8665 & $\sim 43h$ \\
& $\checkmark$ & 0.7674 & 0.4970 & 0.8076 & 0.8702 & $\sim 11h$ \\
\specialrule{1pt}{0pt}{0pt}
\end{tabular}
\begin{tablenotes}[flushleft]
\item
The models are trained with 2048 anchors using closed-loop data samples.
Both the \replaced{closed-loop matching window}{posterior planning horizon} and positive matching horizon are 0.5s. 
The metrics are computed on the validation subset of 150 scenes.
\end{tablenotes}
\end{threeparttable}
}
\label{table:approximate_posterior_policy}
\vspace{-11pt}
\end{table}

\subsubsection{Continuous Regression} \label{Exp: Continuous Regression}

\begin{table}[tb]
\setlength\tabcolsep{2pt}
\centering
\caption{
WOSAC Metrics of Anchor-Based Models With and Without Continuous Regression on a Subset of WOMD Validation Set
}
\resizebox{0.99\columnwidth}{!}{
\def\arraystretch{1.1}
\begin{threeparttable}
\begin{tabular}{cc|cccc|c}
\specialrule{1pt}{0pt}{0pt}
\multirow{2}{*}{\makecell{Closed-\\Loop}} & \multirow{2}{*}{\makecell{Continuous\\Regression}} & \multirow{2}{*}{\textbf{Meta}} & \multirow{2}{*}{Kinematic} & \multirow{2}{*}{Interactive} & \multirow{2}{*}{Map-based} & \multirow{2}{*}{\makecell{Training\\Time}} \\ & & & & & & \\
\hline
\multirow{2}{*}{$\times$}
& $\times$ & 0.6666 & 0.4190 & 0.7269 & 0.7306 & $\sim 10h$ \\
& $\checkmark$ & 0.7361 & 0.4354 & 0.7902 & 0.8384 & $\sim 10h$ \\
\hline
\multirow{2}{*}{$\checkmark$}
& $\times$ & 0.7591 & 0.4921 & 0.8056 & 0.8518 & $\sim 11h$ \\
& $\checkmark$ & 0.7674 & 0.4970 & 0.8076 & 0.8702 & $\sim 11h$ \\
\specialrule{1pt}{0pt}{0pt}
\end{tabular}
\begin{tablenotes}[flushleft]
\item
The models are trained with 2048 anchors and a 0.5s prediction horizon.
The model with continuous regression utilizes the approximate \replaced{closed-loop data generation}{posterior} policy to generate closed-loop samples.
\end{tablenotes}
\end{threeparttable}
}
\label{table:continuous_regression}
\vspace{-11pt}
\end{table}

For the anchor-based model using 2048 anchors and a 0.5s prediction horizon, training it without continuous regression on closed-loop samples actually aligns with the \replaced{discrete NTP}{GPT-like discrete} models~\cite{wu2024smart}.
As shown in Table~\ref{table:continuous_regression}, such a model is indeed competitive.
Building on the consistent configurations, we introduce continuous regression and observe a performance gain.
Moreover, when training with open-loop samples, the model with discrete distributions shows a significant drop in performance, which continuous regression can effectively alleviate.
The results above indicate \added{the key points below}:
\added{(1)}~\textit{The success of discrete \replaced{NTP}{mixture} models hinges on the use of closed-loop data samples}\replaced{;}{.}
\added{(2)}~\textit{The additional flexibility brought by continuous regression is beneficial}.

Thanks to the architectural design in Section~\ref{Motion Decoder}, continuous regression does not incur notable extra computational overhead~(Table~\replaced{X}{\ref*{table:consumption}} \added{in Appendix~\replaced{C-B}{\ref*{appendix:consumption}}}).
Furthermore, if the anchor-based model adopts the approximate \replaced{closed-loop data generation}{posterior} policy~(Section~\ref{Approximate Posterior Policy}), generating closed-loop samples with or without continuous regression is exactly the same.
Thus, \textit{the additional continuous regression does not necessarily demand significantly more training time}, as shown in Table~\ref{table:continuous_regression}.

\subsubsection{\added{Training-Time Overhead}} \label{Training-Time Overhead}

\begin{figure}[tb]
\centering
\includegraphics[width=0.99\columnwidth]{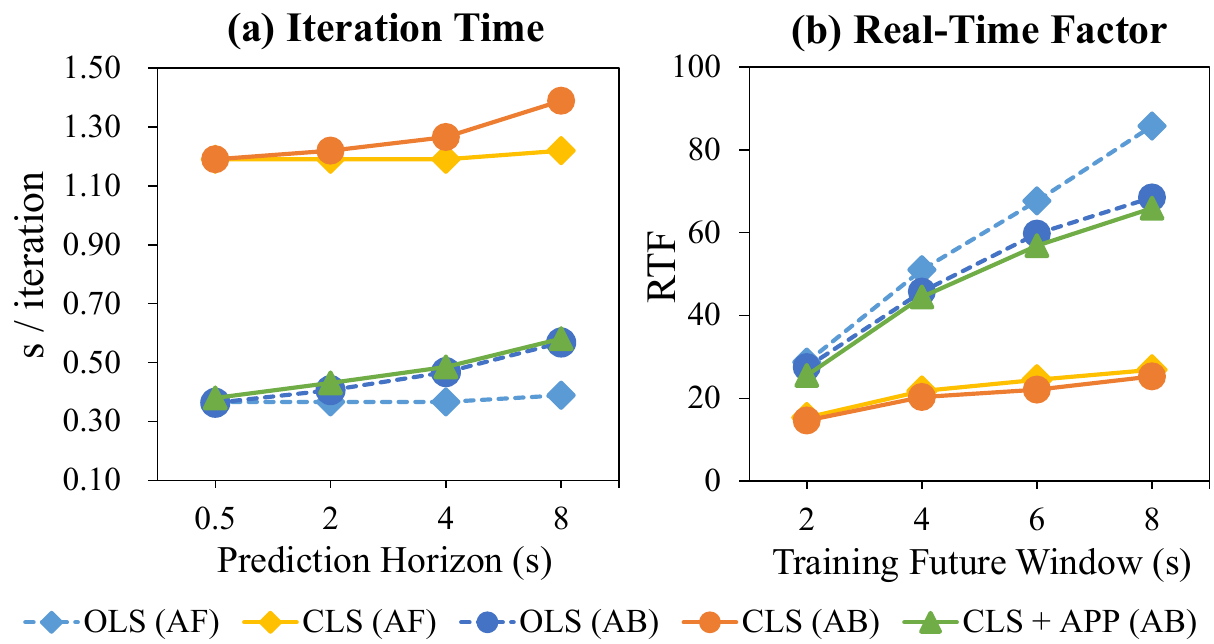}
\caption{\added{
Training-time overhead with a batch size of 4 on a single RTX 3090.
Real-time factor = (batch size × training future window) / iteration time.
``OLS'': open-loop samples; ``CLS'': closed-loop samples; ``APP'': approximate closed-loop data generation policy; ``AF'': anchor-free; ``AB'': anchor-based.
AF models use 6 components; AB models use 2048.
The default prediction horizon and training future window are respectively 4s and 8s.
}}
\label{fig:training_overhead}
\vspace{-11pt}
\end{figure}

\added{
We quantify the efficiency impact of training with closed-loop samples.
As shown in Fig.~\ref{fig:training_overhead}(a), iteration time increases moderately with the prediction horizon, whereas vanilla closed-loop introduces a substantial per-iteration overhead across all horizons.
Anchor-based models are more sensitive to the horizon, primarily due to the need for distance computation between numerous anchors and the ground truth.
In Fig.~\ref{fig:training_overhead}(b), enlarging the training future window proportionally increases the number of rolling steps for closed-loop training, causing its real-time factor (RTF) to plateau.
In contrast, open-loop training benefits from temporal parallelism, achieving nearly linear RTF scaling.
For anchor-based models, employing the approximate closed-loop data generation policy avoids iterative model inference during closed-loop sample generation, reducing the overhead to a level comparable to open-loop training.
}

\subsection{\added{Key Designs for Realistic Simulation}} \label{Key Designs for Realistic Simulation}

\begin{table}[tb]
\setlength\tabcolsep{3pt}
\centering
\caption{
\added{WOSAC Realism Meta Scores of Representative\\Configuration Settings}
}
\resizebox{0.99\columnwidth}{!}{
\def\arraystretch{1.1}
\begin{threeparttable}
\begin{tabular}{ccccc|c}
\specialrule{1pt}{0pt}{0pt}
\multirow{2}{*}{\makecell{Anchor-\\Based}} & \multirow{2}{*}{\makecell{Number of\\Components}} & \multirow{2}{*}{\makecell{Prediction\\Horizon}} & \multirow{2}{*}{\makecell{Continuous\\Regression}} & \multirow{2}{*}{\makecell{Closed-\\Loop}} & \multirow{2}{*}{Realism $\uparrow$} \\ & & & & & \\
\hline
$\times$ & 6 & $4s$ & $\checkmark$ & $\times$ & 0.7409 \\
$\times$ & 1 & $0.5s=\tau$ & $\checkmark$ & $\checkmark$ & 0.7081 \\
$\times$ & 6 & $0.5s=\tau$ & $\checkmark$ & $\checkmark$ & 0.7404 \\
\rowcolor{green!10}
$\times$ & 6 & $4s>\tau$ & $\checkmark$ & $\checkmark$ & \textbf{0.7624} \\
\hline
\checkmark & 64 & $4s$ & $\checkmark$ & $\times$ & 0.7388 \\
\checkmark & 2048 & $4s$ & $\checkmark$ & $\times$ & 0.7570 \\
\checkmark & 2048 & $0.5s=\tau$ & $\times$ & $\checkmark$ & 0.7591 \\
\checkmark & 2048 & $0.5s=\tau$ & $\checkmark$ & $\checkmark$ & 0.7674 \\
\rowcolor{green!10}
\checkmark & 2048 & $4s>\tau$ & $\checkmark$ & $\checkmark$ & \textbf{0.7686} \\
\specialrule{1pt}{0pt}{0pt}
\end{tabular}
\end{threeparttable}
}
\label{table:exploration_highlights}
\vspace{-11pt}
\end{table}

\added{
Within the unified mixture model~(UniMM) framework, our experiments and analyses in Sections~\ref{Training with Open-Loop Samples} and~\ref{Training with Closed-Loop Samples} culminate in new solutions~(highlighted in Table~\ref{table:exploration_highlights}) that will be shown to achieve state-of-the-art realism in Section~\ref{Benchmark Results}.
In the following, we summarize the key designs that underpin these solutions, focusing on the two main challenges: multimodality and distributional shifts.
}

\subsubsection{\added{Multimodality}}

\added{
To support higher multimodality capacity, we design the unified motion decoder for anchor-based models~(Section~\ref{Motion Decoder}) that allows regression-based variants to efficiently scale up the number of components to an unprecedented level, resulting in significantly improved simulation realism~(row~5~$\to$~6 in Table~\ref{table:exploration_highlights}).
}

\added{
Previously, closed-loop data configuration~\cite{venkatraman2015improving, wu2024smart} was primarily applied to unimodal, single-step predictors~(row~2 in Table~\ref{table:exploration_highlights}) or discrete models~(row~7).
In contrast, the closed-loop sample generation method proposed in UniMM~(Section~\ref{Closed-Loop Sample Generation}) enables models under closed-loop training to leverage multiple components~(row~2~$\to$~3) and continuous regression~(row~7~$\to$~8) to better capture multimodal behaviors.
}

\subsubsection{\added{Distributional Shifts}}

\added{
Through a systematic study within UniMM, we identify closed-loop data configuration as a crucial factor in mitigating distributional shifts and enhancing simulation realism.
To generalize its effectiveness, we attempt to apply closed-loop data to general mixture models, and find that long prediction horizons---while beneficial under open-loop training---give rise to the shortcut learning and off-policy learning issues in closed-loop settings.
We address these challenges by disentangling and aligning the relevant time horizons~(Section~\ref{Disentanglement and Alignment of Horizons}), which enables closed-loop samples to effectively benefit regression-based mixture models with long prediction horizons, across both anchor-free~(row~1,~3~$\to$~4 in Table~\ref{table:exploration_highlights}) and anchor-based~(row~6,~8~$\to$~9) paradigms, leading to superior simulation realism.
}

\subsection{Benchmark Results} \label{Benchmark Results}
Based on the exploratory experiments in Sections~\ref{Training with Open-Loop Samples} and \ref{Training with Closed-Loop Samples}, we select the following distinct configurations within the unified mixture model~(\textbf{UniMM}) framework for evaluation on the WOSAC~\cite{montali2024waymo} benchmark:
\begin{itemize}[leftmargin=1em]
\item \textbf{UniMM~(Discrete):}
The anchor-based model \textit{without} continuous regression, using $K=2048$ anchors and a $T_{\text{pred}}=0.5s$ prediction horizon, trained with \textit{closed-loop} samples.
This configuration aligns with the \replaced{discrete NTP models}{GPT-like models utilizing discrete distributions}~\cite{wu2024smart}.
\item \textbf{UniMM~(Anchor-Free):}
The anchor-free model with $K=6$ components and a $T_{\text{pred}}=4s$ prediction horizon, trained with \textit{closed-loop} samples based on a \replaced{$T_{\text{cl}}=0.5s$ closed-loop matching window}{$T_{\text{post}}=0.5s$ posterior planning horizon}.
The horizon for positive component matching, $T_{z^*}=4s$, aligns with the prediction horizon $T_{\text{pred}}$.
This configuration exhibits the best performance among anchor-free models.
\item \textbf{UniMM~(Anchor-Based-0.5s):}
The anchor-based model \textit{with} continuous regression, using $K=2048$ anchors and a $T_{\text{pred}}=\mathbf{0.5s}$ prediction horizon, trained with \textit{closed-loop} samples based on the \textit{approximate} \replaced{closed-loop data generation}{posterior} policy with a \replaced{$T_{\text{cl}}=0.5s$ matching window}{$T_{\text{post}}=0.5s$ planning horizon}.
This configuration is one of the best-performing among anchor-based models and is \textit{as efficient as the discrete model}.
\item \textbf{UniMM~(Anchor-Based-4s):}
The anchor-based model \textit{with} continuous regression, using $K=2048$ anchors and a $T_{\text{pred}}=\mathbf{4s}$ prediction horizon, trained with \textit{closed-loop} samples based on the \textit{approximate} \replaced{closed-loop data generation }{posterior }policy with a \replaced{$T_{\text{cl}}=0.5s$ matching window}{$T_{\text{post}}=0.5s$ planning horizon}.
The horizon for positive component matching, $T_{z^*}=0.5s$, aligns with the \replaced{closed-loop matching window $T_{\text{cl}}$}{posterior planning horizon $T_{\text{post}}$}.
This configuration is one of the best-performing among anchor-based models.
\end{itemize}

\begin{table*}[tb]
\centering
\caption{
WOSAC Benchmark Results on the Validation and Testing Sets of WOMD
}
\resizebox{0.95\textwidth}{!}{
\def\arraystretch{1.2}
\begin{tabular}{c|lr|ccccc}
\specialrule{1pt}{0pt}{0pt}
Set & Method & Params & \textbf{Realism Meta} $\uparrow$ & Kinematic $\uparrow$ & Interactive $\uparrow$ & Map-based $\uparrow$ & minADE $\downarrow$ \\
\hline
\multirow{8}{*}{Test}
& VBD~\cite{huang2024versatile} & 12M & 0.7200 & 0.4169 & 0.7819 & 0.8137 & 1.4743 \\
& MVTE~\cite{wang2023multiverse} & - & 0.7302 & 0.4503 & 0.7706 & 0.8381 & 1.6770 \\
& MPS~\cite{lou2024model} & 65M & 0.7417 & 0.4182 & 0.7942 & 0.8591 & 1.4842 \\
& GUMP~\cite{hu2025solving} & 523M & 0.7431 & 0.4780 & 0.7887 & 0.8359 & 1.6041 \\
& BehaviorGPT~\cite{zhou2024behaviorgpt} & 3M & 0.7473 & 0.4333 & 0.7997 & 0.8593 & 1.4147 \\
& KiGRAS~\cite{zhao2024kigras} & 0.7M & 0.7597 & 0.4691 & 0.8064 & 0.8658 & 1.4383 \\
& SMART~\cite{wu2024smart} & 7M & 0.7591 & 0.4759 & 0.8039 & 0.8632 & 1.4062 \\
& SMART~\cite{wu2024smart} & 101M & 0.7614 & 0.4786 & 0.8066 & 0.8648 & 1.3728 \\
& \textbf{UniMM (Discrete)} & 4M & 0.7621 & 0.4877 & 0.8052 & 0.8634 & 1.3383 \\
& \textbf{UniMM (Anchor-Free)} & 4M & 0.7628 & 0.4746 & 0.8080 & 0.8693 & 1.3154 \\
& \textbf{UniMM (Anchor-Based-0.5s)} & 4M & 0.7675 & \textbf{0.4934} & 0.8082 & 0.8718 & 1.3300 \\
& \textbf{UniMM (Anchor-Based-4s)} & 4M & \textbf{0.7684} & 0.4913 & \textbf{0.8101} & \textbf{0.8730} & \textbf{1.2947} \\
\hline
\multirow{3}{*}{Val}
& \textbf{UniMM (Discrete)} & 4M & 0.7618 & 0.4875 & 0.8051 & 0.8629 & 1.3473 \\
& \textbf{UniMM (Anchor-Free)} & 4M & 0.7631 & 0.4744 & 0.8083 & 0.8700 & 1.3208 \\
& \textbf{UniMM (Anchor-Based-0.5s)} & 4M & 0.7674 & \textbf{0.4929} & 0.8083 & 0.8718 & 1.3403 \\
& \textbf{UniMM (Anchor-Based-4s)} & 4M & \textbf{0.7687} & 0.4912 & \textbf{0.8105} & \textbf{0.8737} & \textbf{1.2980} \\
\specialrule{1pt}{0pt}{0pt}
\end{tabular}
}
\label{table:benchmark}
\vspace{-8pt}
\end{table*}

As shown in Table~\ref{table:benchmark}, \textbf{UniMM~(Discrete)} is comparable to current state-of-the-art models utilizing discrete distributions~\cite{wu2024smart,zhao2024kigras}, suggesting that the analysis within the unified mixture model framework is sound.
The highly competitive performance of \textbf{UniMM~(Anchor-Free)} clearly demonstrates \added{the following}:
\added{(1)~For \textbf{Q1},} \textit{\replaced{model}{Model} configurations cannot fully explain the performance gap between \replaced{existing regression-based mixture models and discrete NTP models}{continuous and discrete mixture models}\deleted{~(\textbf{Q1})}}\replaced{;}{.}
\added{(2)~For \textbf{Q2},} \textit{\replaced{the}{The} key to superior performance lies in closed-loop samples, enabling model configurations distinct from \replaced{discrete NTP}{GPT-like discrete} models to achieve realistic simulations as well\deleted{~(\textbf{Q2})}}.
Additionally, due to the flexible predictions of each component, the anchor-free model with several mixture components can effectively generate multimodal agent behaviors.
\textbf{UniMM~(Anchor-Based-4s)} achieves the best simulation realism, further \replaced{addressing \textbf{Q3}}{indicating that}:
\textit{By addressing the shortcut learning issue and the off-policy learning problem, the use of closed-loop samples can benefit general mixture models, particularly those with a longer prediction horizon\deleted{~(\textbf{Q3})}}.
Moreover, with efficiency comparable to its discrete counterpart, \textbf{UniMM~(Anchor-Based-0.5s)} exhibits superior effectiveness, which highlights the advantages of continuous modeling in multi-agent simulation.

\subsection{Qualitative Results} \label{Qualitative Results}
\begin{figure*}[tb]
\centering
\setlength{\abovecaptionskip}{-2pt}
\includegraphics[width=0.93\textwidth]{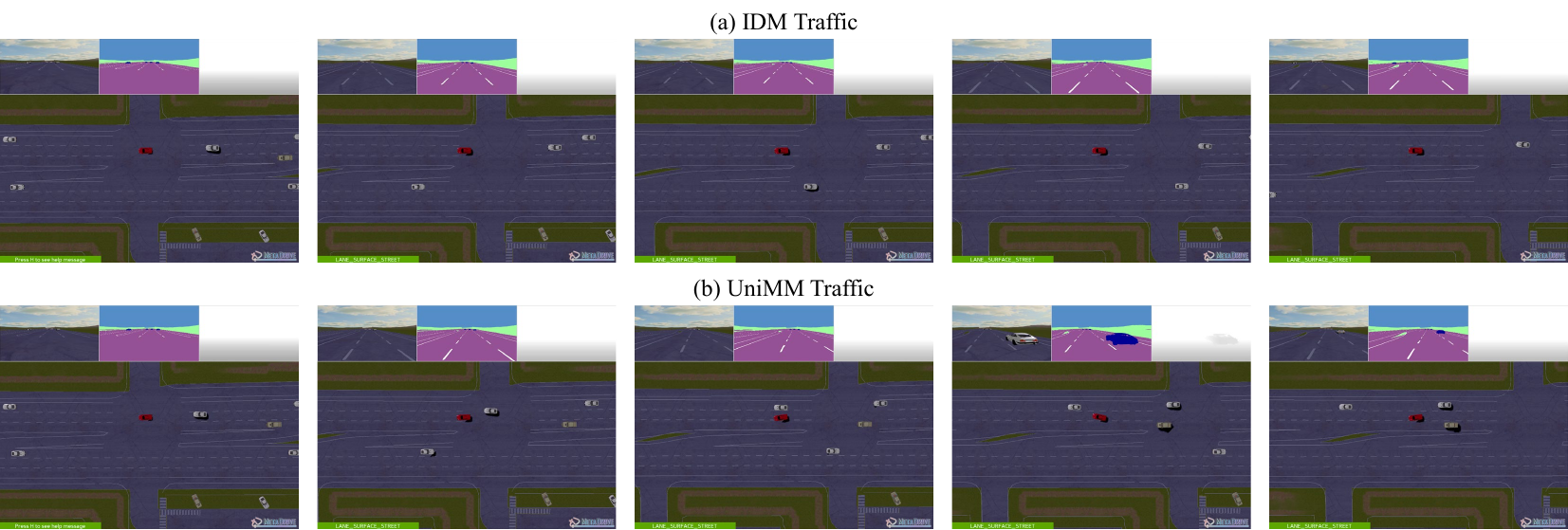}
\caption{\added{
Scene visualizations under different background traffic models in the MetaDrive simulator.
The red vehicle denotes the ego agent controlled by the PPO policy; other traffic agents are governed by (a) IDM or (b) UniMM.
The three top panels of each frame respectively show the MetaDrive-rendered RGB, semantic, and point-cloud views.
}}
\label{fig:metadrive}

\setlength{\abovecaptionskip}{-5pt}
\includegraphics[width=0.99\textwidth]{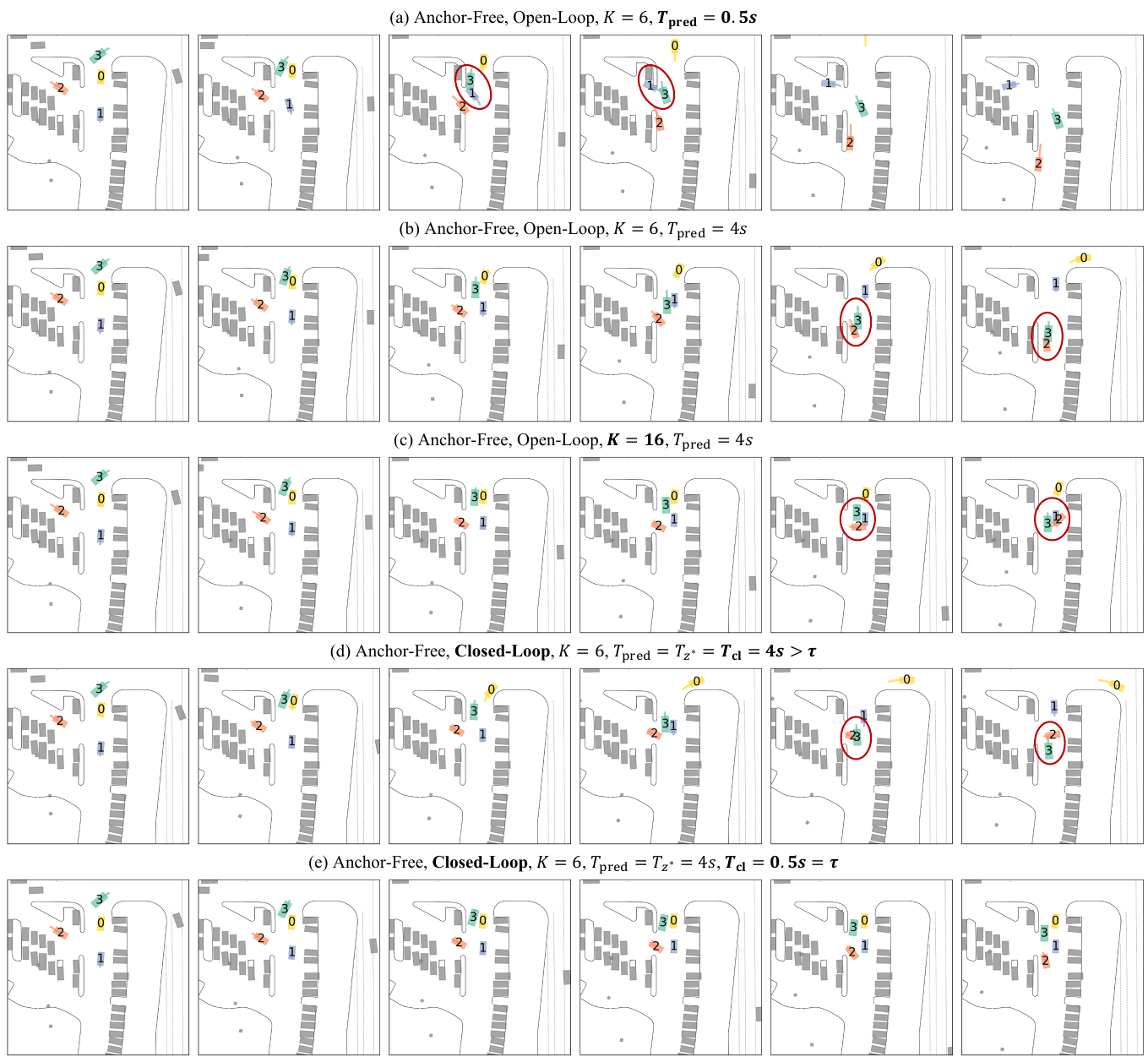}
\caption{
Qualitative results of anchor-free models.
``Open/Closed-Loop'' indicates that the model is trained using open-loop or closed-loop samples.
$K$, $T_{\text{pred}}$, $T_{z^*}$, \replaced{$T_{\text{cl}}$}{$T_{\text{post}}$}, and $\tau$ respectively denote the number of components, prediction horizon, positive matching horizon, \replaced{closed-loop matching window}{posterior planning horizon}, and simulation update interval.
}
\label{fig:qualitative_results_anchor-free}
\end{figure*}

\begin{figure*}[tb]
\centering
\includegraphics[width=0.99\textwidth]{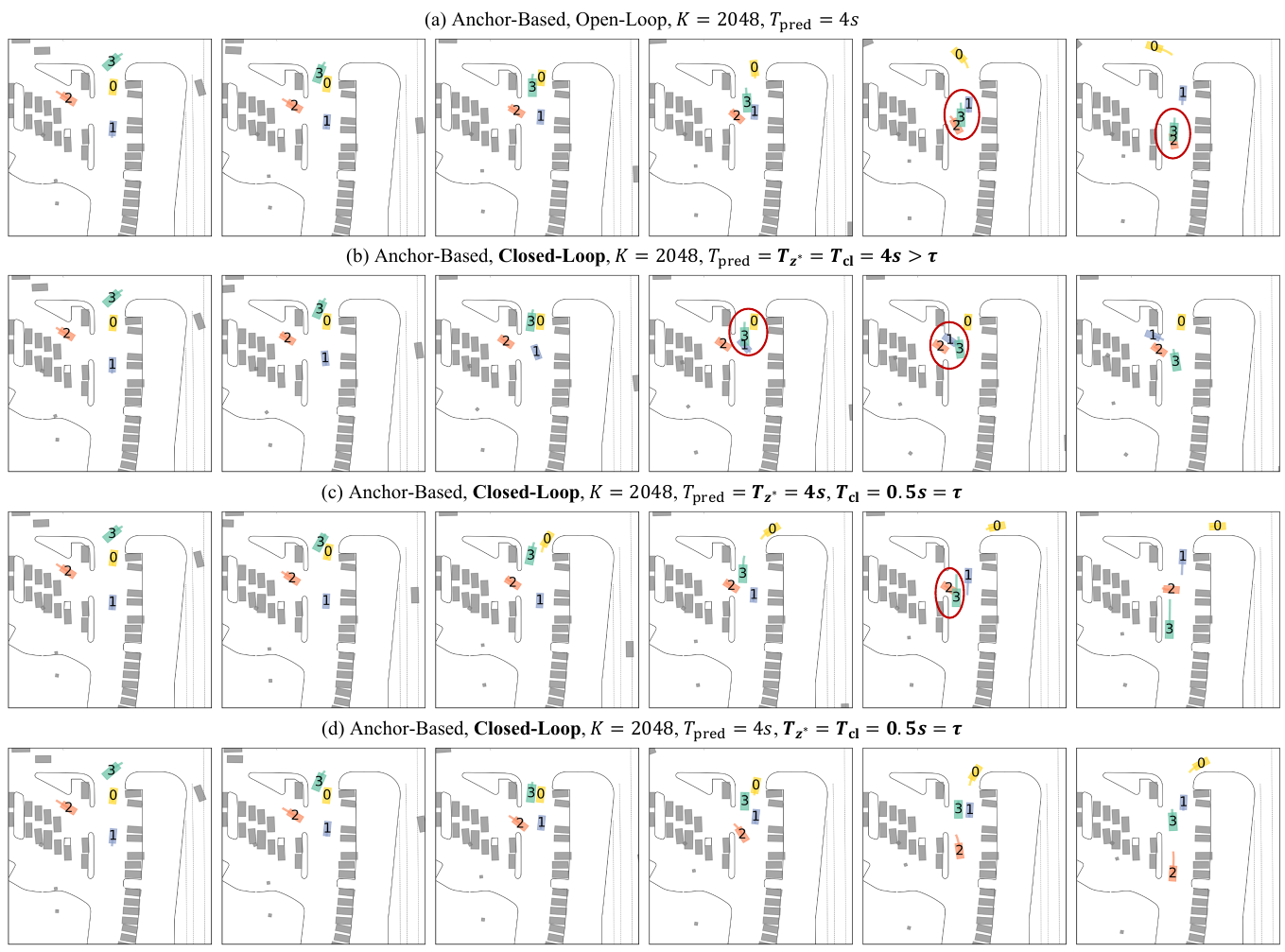}
\caption{
Qualitative results of anchor-based models \textit{with} continuous regression.
``Open/Closed-Loop'' indicates that the model is trained using open-loop or closed-loop samples.
$K$, $T_{\text{pred}}$, $T_{z^*}$, \replaced{$T_{\text{cl}}$}{$T_{\text{post}}$}, and $\tau$ respectively denote the number of components, prediction horizon, positive matching horizon, \replaced{closed-loop matching window}{posterior planning horizon}, and simulation update interval.
}
\label{fig:qualitative_results_anchor-based}

\vspace{10pt}

\includegraphics[width=0.99\textwidth]{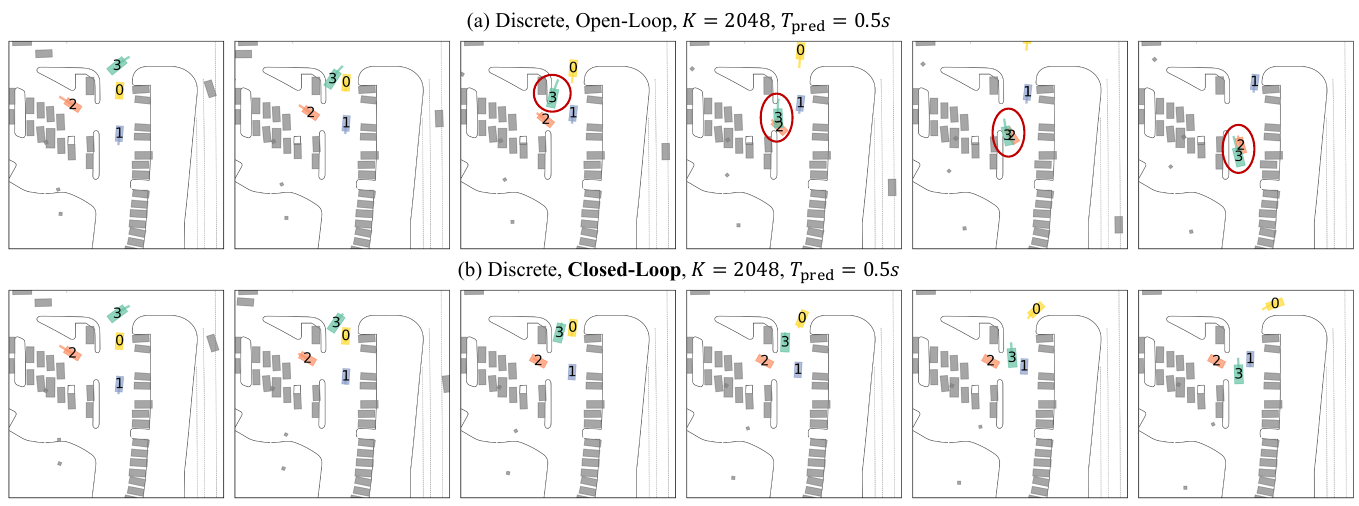}
\caption{
Qualitative results of anchor-based models \textit{without} continuous regression.
``Open/Closed-Loop'' indicates that the model is trained using open-loop or closed-loop samples.
$K$ and $T_{\text{pred}}$ respectively denote the number of components and prediction horizon.
}
\label{fig:qualitative_results_discrete}
\end{figure*}

To intuitively grasp the characteristics of different configurations, we plot the representative scenarios generated by various models in Fig.~\ref{fig:qualitative_results_anchor-free}, Fig.~\ref{fig:qualitative_results_anchor-based}, and Fig.~\ref{fig:qualitative_results_discrete}.

As shown in Fig.~\ref{fig:qualitative_results_anchor-free}(a), the model with a short prediction horizon often generates highly unrealistic behaviors, such as vehicles crossing each other or veering off the road.
In contrast, a longer prediction horizon promotes robust behavior, but training with open-loop samples could still result in out-of-distribution~(OOD) scenarios, like the rear-end collision in Fig.~\ref{fig:qualitative_results_anchor-free}(b).
Additionally, Fig.~\ref{fig:qualitative_results_anchor-free}(c) shows that, for anchor-free models, a larger number of components may lead to the selection of more unrealistic trajectories, resulting in implausible scenarios such as vehicle overlapping.

Fig.~\ref{fig:qualitative_results_anchor-free}(b), Fig.~\ref{fig:qualitative_results_anchor-based}(a), and Fig.~\ref{fig:qualitative_results_discrete}(a) indicate that models trained with open-loop data samples, across various model configurations, are prone to generate OOD scenarios, such as collisions or off-road driving.
Furthermore, Fig.\ref{fig:qualitative_results_anchor-free}(d) and Fig.\ref{fig:qualitative_results_anchor-based}(b) illustrate the adverse effects of the shortcut learning issue, while Fig.~\ref{fig:qualitative_results_anchor-based}(c) highlights the negative impact of the off-policy learning problem.
Finally, Fig.~\ref{fig:qualitative_results_anchor-free}(e), Fig.~\ref{fig:qualitative_results_anchor-based}(d), and Fig.~\ref{fig:qualitative_results_discrete}(b) demonstrate that closed-loop samples facilitate various mixture models in generating realistic simulations, provided that the shortcut learning and off-policy learning issues are effectively addressed.

\subsection{\added{Toward Real-World System Application}} \label{Toward Real-World System Application}

\begin{table}[tb]
\setlength\tabcolsep{2pt}
\centering
\caption{
\added{Ego Policy Evaluation under Different Background Traffic Models in the MetaDrive Simulator}
}
\resizebox{0.99\columnwidth}{!}{
\def\arraystretch{1.1}
\begin{threeparttable}
\begin{tabular}{c|c|cccc|c}
\specialrule{1pt}{0pt}{0pt}
\multirow{2}{*}{Traffic} & \multirow{2}{*}{\makecell{Ego\\Policy}} & \multirow{2}{*}{Success} & \multirow{2}{*}{Collision} & \multirow{2}{*}{Offroad} & \multirow{2}{*}{Progress} & \multirow{2}{*}{\makecell{Traffic\\Speed}} \\ & & & & & & \\
\hline
\multirow{2}{*}{IDM}
& IDM & 0.66 & 0.03 & 0.00 & 0.86 & 3.94 m/s \\
& PPO & 0.40 & 0.10 & 0.00 & 0.73 & 3.96 m/s \\
\hline
\multirow{2}{*}{UniMM}
& IDM & 0.60 & 0.06 & 0.00 & 0.83 & 4.49 m/s \\
& PPO & 0.41 & 0.16 & 0.00 & 0.74 & 4.50 m/s \\
\specialrule{1pt}{0pt}{0pt}
\end{tabular}
\begin{tablenotes}[flushleft]
\scriptsize
\item
The evaluation is conducted on 100 scenarios.
``Success'', ``Collision'', and ``Offroad'' are episode-level rates.
Success requires completing the route within the time limit without any collision or offroad.
``Progress'' is the route completion ratio at termination.
``Traffic Speed'' is the mean speed of background traffic agents.
\end{tablenotes}
\end{threeparttable}
}
\label{table:metadrive}
\vspace{-11pt}
\end{table}


\added{
Simulation-based evaluation is a common step for real-world autonomous driving system development.
We integrate UniMM as a background traffic model into MetaDrive~\cite{li2022metadrive}, a simulator that provides accurate physics and realistic sensor rendering~(Fig.~\ref{fig:metadrive}).
Specifically, UniMM iteratively predicts the future states of all agents based on the simulator-maintained scene states, while only using the predictions to update the non-ego traffic agents.
}

\added{
In MetaDrive~\cite{li2022metadrive}, we evaluate the built-in ego policies~(IDM and LiDAR-based PPO) under the UniMM background traffic, compared to the default IDM background traffic.
As shown in Table~\ref{table:metadrive}, UniMM traffic yields overall evaluation trends similar to IDM traffic, providing a reasonable assessment of ego policies.
Nevertheless, unlike the conservative IDM, UniMM produces higher traffic speeds and can generate overtaking maneuvers~(Fig.~\ref{fig:metadrive}), suggesting that it captures more assertive human driving behaviors and introduces richer interactions for the ego policy.
We also observe policy-dependent differences in specific metrics.
For example, under UniMM traffic, the success rate of IDM policy decreases noticeably, whereas PPO remains largely unchanged.
These results demonstrate that UniMM contributes novel and complementary value to driving assessment.
}

\section{Conclusion}
In this study, we \replaced{formulate a unified mixture model framework~(UniMM)}{revisit mixture models} for multi-agent simulation, seeking to unify mainstream methods, including \replaced{regression-based}{continuous} mixture models and \replaced{discrete NTP}{GPT-like discrete} models, which originate from different inspirations.
Within the \replaced{UniMM}{unified mixture model} framework, we identify critical model configurations \replaced{and comprehensively characterize their effects through a systematic investigation.}{, including positive component matching, continuous regression, prediction horizon, and number of components.
Through a systematic investigation of these model configurations, we uncover more comprehensive trends in their impacts.}
Furthermore, we introduce a closed-loop sample generation approach for general mixture models, highlighting its connection to motion tokenization in \replaced{discrete NTP}{GPT-like discrete} models.
Our experiments reveal that the \added{closed-loop} data configuration\deleted{ incorporating closed-loop samples} is key to achieving realistic simulations.
\replaced{We further uncover the shortcut learning and off-policy learning issues that prevent closed-loop samples from benefiting models with long prediction horizons, and we propose a temporal disentanglement-and-alignment mechanism to address these issues.}{To extend the benefits of closed-loop samples to a broader range of mixture models, we further analyze and address the shortcut learning and off-policy issues.}
Building on our exploration, the distinct variants proposed in the \deleted{unified mixture model~(}UniMM\deleted{)} framework, including discrete, anchor-free, and anchor-based models, achieve state-of-the-art performance on the WOSAC benchmark.

\noindent\textbf{\added{Limitations and broader scope.}}
\added{
This study centers on supervised imitation learning~(IL).
Complementary work refines multi-agent simulation via reinforcement learning~(RL), which promotes reward-aligned behaviors beyond the logged data~\cite{rowe2024ctrl, chen2025rift} and can further mitigate distributional shifts through online rollouts~\cite{zhang2023learning, peng2024improving}.
Within this broader RL-inclusive scope, IL still commonly serves as a pretraining stage or regularization term~\cite{pei2025advancing, peng2024improving}, efficiently distilling offline experience and stabilizing training.
Incorporating both IL and RL into the UniMM framework represents a promising direction for future work.
}

\clearpage



%
\bibliography{main}
\bibliographystyle{IEEEtran}


\vspace{-31pt}
\begin{IEEEbiography}[{\includegraphics[width=1in,height=1.25in,clip,keepaspectratio]{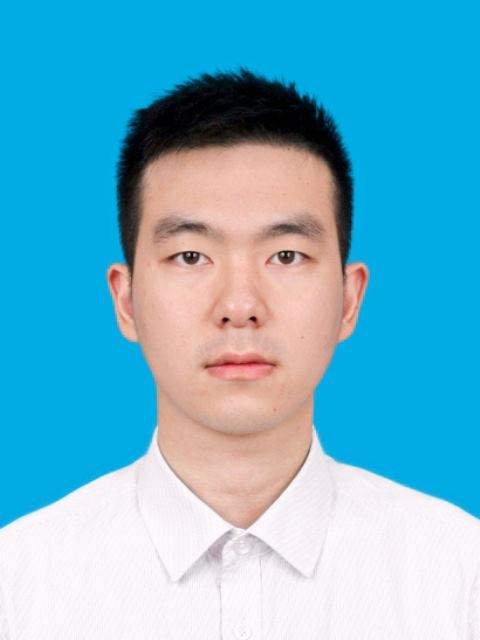}}]{Longzhong Lin}
is a Ph.D. student at Zhejiang University. Before that, he received his B.S. degree from College of  Control Science and Engineering, Zhejiang University, China. His research interests include autonomous robots and robot learning.
\end{IEEEbiography}

\begin{IEEEbiography}[{\includegraphics[width=1in,height=1.25in,clip,keepaspectratio]{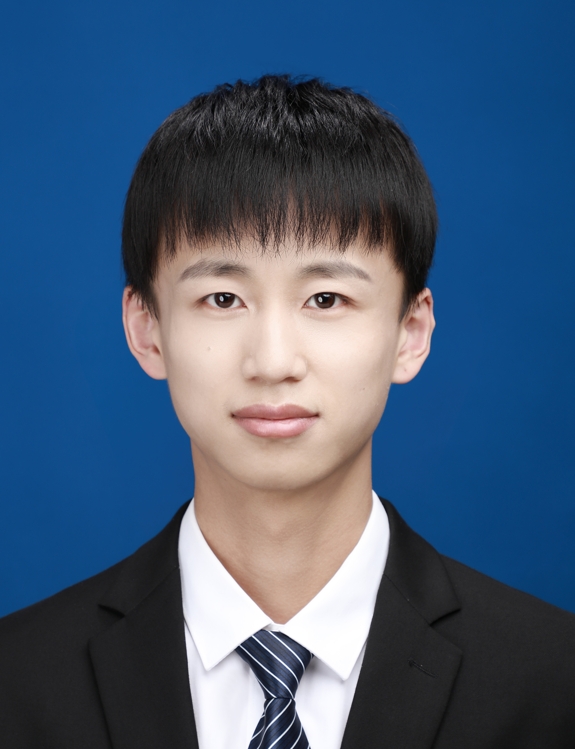}}]{Xuewu Lin}
received the Master's degree from the School of Vehicle and Mobility, Tsinghua University, Beijing, China, in 2021. He is currently working as a Researcher at Horizon Robotics Inc. His current research interests include embodied intelligence and autonomous driving.
\end{IEEEbiography}
\vspace{-13pt}

\begin{IEEEbiography}[{\includegraphics[width=1in,height=1.25in,clip,keepaspectratio]{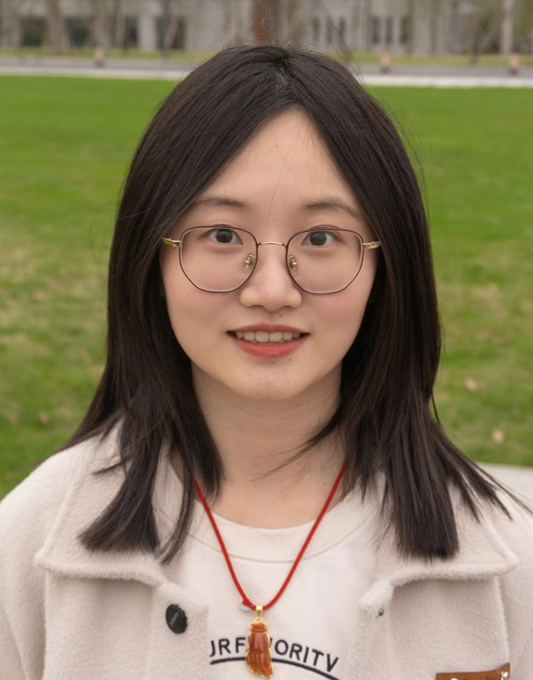}}]{Kechun Xu}
received her B.Eng. in Control Science and Engineering from Zhejiang University, Hangzhou, China, in 2021. She is currently working toward Ph.D. degree at the State Key Laboratory of Industrial Control Technology and Institute of Cyber-Systems and Control, Zhejiang University, Hangzhou, China. Her research interests include manipulation and robot learning.
\end{IEEEbiography}
\vspace{-13pt}

\begin{IEEEbiography}[{\includegraphics[width=1in,height=1.25in,clip,keepaspectratio]{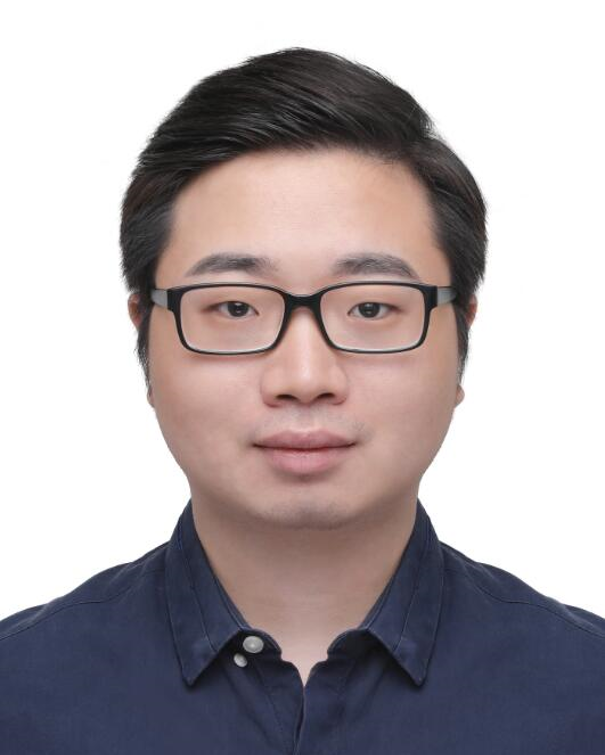}}]{Haojian Lu (Member, IEEE)}
received Ph.D. degree in Robotics from City University of Hong Kong in 2019. He was a Research Assistant at City University of Hong Kong, from 2019 to 2020. He is currently a professor in the Department of Control Science and Engineering, Zhejiang University. His research interests include bioinspired robotics, medical robotics, and soft robotics.
\end{IEEEbiography}
\vspace{-13pt}

\begin{IEEEbiography}[{\includegraphics[width=1in,height=1.25in,clip,keepaspectratio]{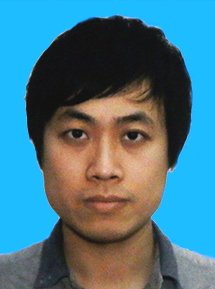}}]{Lichao Huang}
received the Master's degree from the Department of Computing, Imperial College, London in 2014. He is currently working as a Researcher at Horizon Robotics Inc. His research interests include deep learning for computer vision and autonomous robots.
\end{IEEEbiography}
\vspace{-13pt}

\begin{IEEEbiography}[{\includegraphics[width=1in,height=1.25in,clip,keepaspectratio]{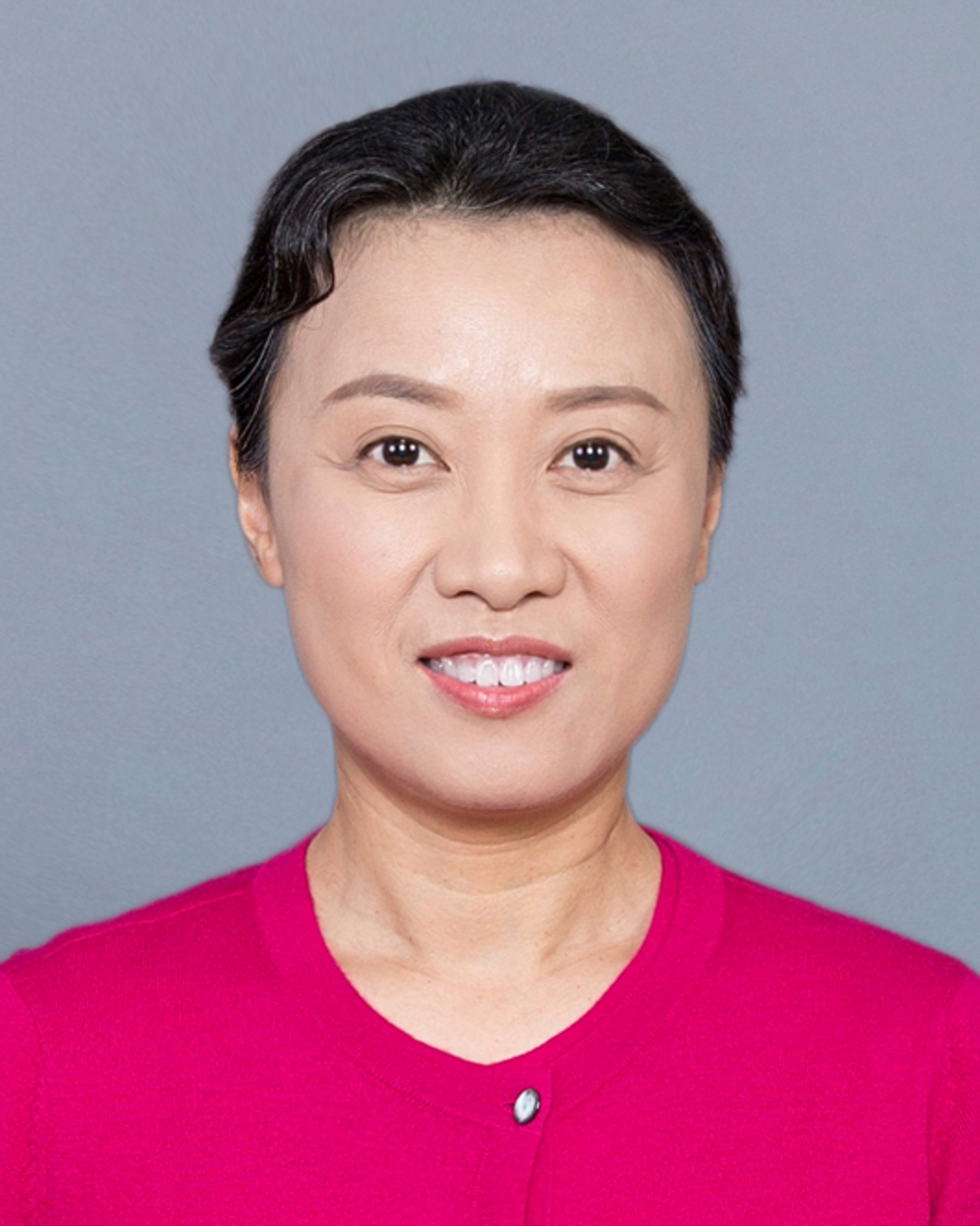}}]{Rong Xiong}
received her Ph.D. in Control Science and Engineering from the Department of Control Science and Engineering, Zhejiang University, China in 2009. She is currently a Professor in the Department of Control Science and Engineering, Zhejiang University, China. Her latest research interests include embodied intelligence and motion control.
\end{IEEEbiography}
\vspace{-13pt}

\begin{IEEEbiography}[{\includegraphics[width=1in,height=1.25in,clip,keepaspectratio]{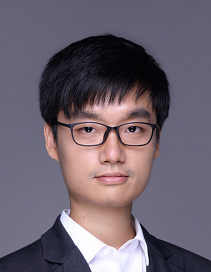}}]{Yue Wang}
received the Ph.D. degree from the Department of Control Science and Engineering, Zhejiang University, Hangzhou, China, in 2016. He is currently working as a Professor with the Department of Control Science and Engineering, Zhejiang University. His current research interests include autonomous robots and robot learning.
\end{IEEEbiography}


\clearpage
\input{appendix}

\end{document}

%% file: appendix.tex
\begin{appendices}

\section{Metrics} \label{appendix:metrics}

\subsection{Multi-Agent Simulation}
For multi-agent simulation, we adopt the evaluation metrics from the Waymo Open Sim Agents Challenge~(WOSAC)~\cite{montali2024waymo}, which include Kinematic metrics~(likelihood of linear speed and acceleration, angular speed and acceleration), Interactive metrics~(likelihood of distance to nearest object, collision, time to collision), and Map-based metrics~(likelihood of distance to road edge, offroad).
The Realism Meta metric serves as an overall measure, derived from a weighted combination of the above metrics.

The WOSAC metrics are by default based on 32 rollouts, each lasting 8 seconds at 10 Hz, for every scenario with 1 second of history from the WOMD validation or testing set.
As the number of scenes is large and the computation of metrics is expensive, evaluating on the full set via the official website is time-consuming.
For the exploratory experiments in Section~\ref{Training with Open-Loop Samples} and Section~\ref{Training with Closed-Loop Samples}, we evaluate models on a subset of the validation set, using a local evaluation tool based on the official API to compute the metrics.
The subset consists of the first scene from each shard of the dataset, containing 150 scenes in total.
The benchmark results in Section~\ref{Benchmark Results} are statistics on the full validation and testing sets, obtained from the official evaluation server.

\subsection{Open-Loop Motion Prediction}
In some cases, we also assess the open-loop prediction performance as a reference, using metrics from the Waymo Motion Prediction Challenge~\cite{ettinger2021large}, including minADE~(Minimum Average Displacement Error), minFDE~(Minimum Final Displacement Error), MR~(Miss Rate), and mAP~(Mean Average Precision).
The motion prediction metrics are computed on multiple 8-second future trajectories of interested agents, given 1 second of history.
We apply the official evaluation tool locally to calculate the metrics on the full WOMD validation set, since the evaluation server only accepts submissions with up to 6 trajectories for each target agent.

\section{Implementation Details} \label{appendix:details}

\subsection{Architecture Details}
\begin{figure}[tb]
\centering
\includegraphics[width=0.99\columnwidth]{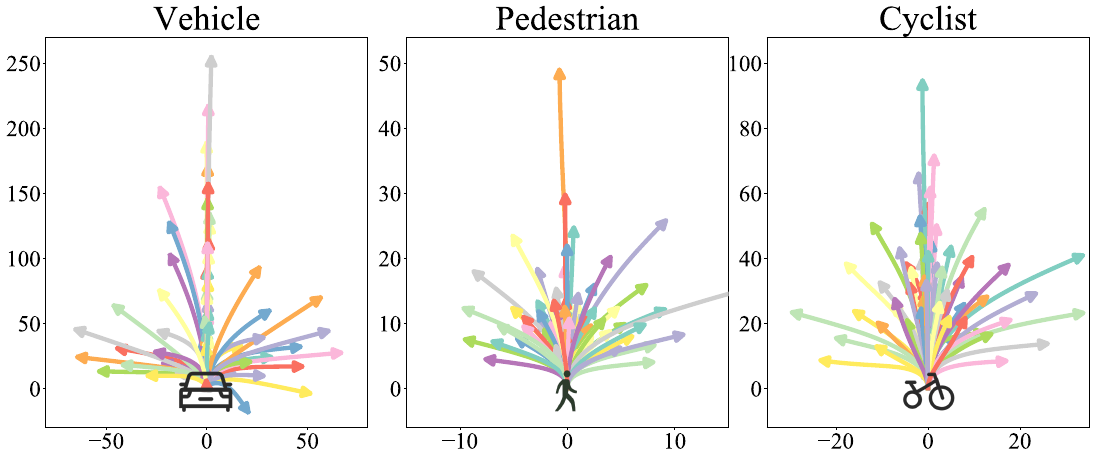}
\caption{
An example of anchor trajectories for each agent category.
These 64 anchor trajectories, each lasting 8 seconds, are generated by k-means clustering on the training set.
}
\label{fig:kmeans_anchors}
\end{figure}

To balance granularity and efficiency, we downsample the HD map so that the point distance is around 2.5 meters and divide it into polylines with intervals of about 5 meters.
The agent trajectories are segmented into multiple tracklets, each with a duration equal to the simulation update interval $\tau = 0.5s$.
The embeddings at the map polyline or agent tracklet level are derived through attention-based aggregation, with the dimensions set to 128.
For modeling interactions among scene elements, we apply 1 layer of map self-attention and stack 2 layers of factorized attention, where the attention operations are performed within a local range.
Specifically, the temporal attention has a time window of 3 seconds, and the numbers of neighbors for map-map, agent-map, and agent-agent attention are 16, 64, and 32, respectively.
The scorer and continuous regression networks in the decoder are implemented using 2-layer MLPs.
In anchor-free models, component queries and agent embeddings are fused by concatenation.
In anchor-based models, anchors are generated through the widely-used k-means clustering on the training set~\cite{chai2019multipath}, as shown in Figure~\ref{fig:kmeans_anchors}.
Before entering the continuous regression network, the selected anchor is encoded by a 2-layer MLP and then concatenated with the agent embedding.
Ultimately, the models in our experiments generally have around 4 million parameters and comparable computational overhead~(Table~\ref{table:consumption}).

\subsection{Inference Details}
The agent state in the model's predicted trajectory includes 2D position and heading.
Following previous works~\cite{shi2022motion,zhou2023query}, we treat each coordinate and time step in the trajectory as independent.
In the component distribution, we use Laplace distributions for position and von Mises distribution for heading.
At inference, the output trajectory for a specific component utilizes the expected value of its corresponding component distribution.
During closed-loop simulation, we sample the components based on the original output probabilities to avoid introducing bias and ensure a fair comparison.

\section{Additional Experiment Results}

\subsection{Motion Prediction Metrics for Anchor-Free Models} \label{appendix:motion_prediction}

\begin{table}[tb]
\centering
\caption{
Motion Prediction Metrics of Anchor-Free Models with Varying Component Numbers
}
\resizebox{0.99\columnwidth}{!}{
\def\arraystretch{1.1}
\begin{threeparttable}
\begin{tabular}{c|cccc}
\specialrule{1pt}{0pt}{0pt}
\multirow{2}{*}{\makecell{Number of\\Components}} & \multirow{2}{*}{mAP $\uparrow$} & \multirow{2}{*}{minADE $\downarrow$} & \multirow{2}{*}{minFDE $\downarrow$} & \multirow{2}{*}{MR $\downarrow$} \\ & & & & \\
\hline
3 & \textbf{0.3356} & 0.8123 & 1.9647 & 0.2635 \\
6 & 0.2877 & 0.7043 & 1.6606 & 0.2088 \\
16 & 0.2037 & \textbf{0.5494} & \textbf{1.1844} & \textbf{0.1377}  \\
16\tnote{*} & 0.1943 & 0.8368 & 2.0180 & 0.2679 \\
\specialrule{1pt}{0pt}{0pt}
\end{tabular}
\begin{tablenotes}[flushleft]
\scriptsize
\item The models are trained with an 8s prediction horizon on open-loop samples.
\item[*] The metrics for the corresponding row are computed based on the top-6 predictions ranked by confidence scores.
\end{tablenotes}
\end{threeparttable}
}
\label{table:anchor-free_pred}
\end{table}

To validate our hypothesis that anchor-free models may struggle to select the appropriate mixture component based on output scores when the number of components is large, we evaluate anchor-free models with varying component numbers on open-loop motion prediction.
As shown in Table~\ref{table:anchor-free_pred}, the mAP, which reflects the ability to score multimodal motions, consistently deteriorates with the increase in component number.
Moreover, the top-6 predictions of the model with 16 components are even worse than the model with 3 components across all prediction metrics.
The above results suggest that a larger number of components may hinder anchor-free models from picking out realistic trajectories.

\subsection{Computational Overhead} \label{appendix:consumption}

\begin{table}[tb]
\setlength\tabcolsep{3pt}
\centering
\caption{
Computational Costs and Inference Latency of Models
}
\resizebox{0.99\columnwidth}{!}{
\def\arraystretch{1.1}
\begin{threeparttable}
\begin{tabular}{cccc|cc}
\specialrule{1pt}{0pt}{0pt}
\multirow{2}{*}{\makecell{Anchor-\\Based}} & \multirow{2}{*}{\makecell{Number of\\Components}} & \multirow{2}{*}{\makecell{Prediction\\Horizon}} & \multirow{2}{*}{\makecell{Continuous\\Regression}} & \multirow{2}{*}{MACs} & \multirow{2}{*}{Latency} \\ & & & & & \\
\hline
\multirow{4}{*}{$\times$}
& 3 & 8s & $\checkmark$ & 20.72G & 33.73ms \\
& 6 & 8s & $\checkmark$ & 21.22G & 33.79ms \\
& 16 & 8s & $\checkmark$ & 22.90G & 33.94ms \\
& 64\tnote{*} & 8s & $\checkmark$ & 30.92G & 34.41ms \\
\hline
\multirow{5}{*}{\checkmark} 
& 64 & 8s & $\checkmark$ & 20.42G & 33.93ms \\
& 512 & 8s & $\checkmark$ & 20.46G & 34.17ms \\
& 2048 & 8s & $\checkmark$ & 20.59G & 35.89ms \\
& 2048 & 0.5s & $\checkmark$ & 20.53G & 33.59ms \\
& 2048 & 0.5s & $\times$ & 20.42G & 33.10ms \\
\specialrule{1pt}{0pt}{0pt}
\end{tabular}
\begin{tablenotes}[flushleft]
\item
\scriptsize
These metrics are averaged over the validation subset of 150 scenes.
\item[*] The corresponding model encounters memory exhaustion during training.
\end{tablenotes}
\end{threeparttable}
}
\label{table:consumption}
\end{table}

We evaluate the computational overhead of UniMM variants under different model configuration settings.
The results are summarized in Table~\ref{table:consumption}, where the computational cost is represented by the MACs required for the model to process a single scene sample, and the latency refers to the time taken by one-step inference for a single scene on an NVIDIA RTX 3090 GPU.

\end{appendices}